\documentclass[letterpaper]{article} 
\usepackage{aaai24}  
\usepackage{times}  
\usepackage{helvet}  
\usepackage{courier}  
\usepackage[hyphens]{url}  
\usepackage{graphicx} 
\usepackage{natbib}  
\usepackage{caption} 
\usepackage{algorithm}
\usepackage{algorithmic}

\usepackage{cancel}
\usepackage{subcaption}
\usepackage{amsmath}
\usepackage{amssymb}
 \usepackage{color}
\usepackage{newfloat}
\usepackage{listings}
\DeclareCaptionStyle{ruled}{labelfont=normalfont,labelsep=colon,strut=off} 
\floatstyle{ruled}
\newfloat{listing}{tb}{lst}{}
\floatname{listing}{Listing}
\title{Make Lossy Compression Meaningful for Low-Light Images}
\author{
    Shilv Cai\textsuperscript{\rm 1,2}, 
    Liqun Chen\textsuperscript{\rm 1,2}, 
    Sheng Zhong\textsuperscript{\rm 1,2}, 
    Luxin Yan\textsuperscript{\rm 1,2}, 
    Jiahuan Zhou\textsuperscript{\rm 3}, 
    Xu Zou\textsuperscript{\rm 1,2,}\thanks{Corresponding author.}
}
\affiliations{
    \textsuperscript{\rm 1}Huazhong University of Science and Technology, Wuhan, Hubei 430074, China\\
    \textsuperscript{\rm 2}National Key Laboratory of Multispectral Information Intelligent Processing Technology, Wuhan, Hubei 430074, China\\
    \textsuperscript{\rm 3}Wangxuan Institute of Computer Technology, Peking University, Beijing 100871, China\\

    \{caishilv, chenliqun, zhongsheng, yanluxin, zoux\}@hust.edu.cn, jiahuanzhou@pku.edu.cn
}

\usepackage{bibentry}
\begin{document}

\maketitle

\begin{abstract}
Low-light images frequently occur due to unavoidable environmental influences or technical limitations, such as insufficient lighting or limited exposure time. 
To achieve better visibility for visual perception, low-light image enhancement is usually adopted. 
Besides, lossy image compression is vital for meeting the requirements of storage and transmission in computer vision applications.
To touch the above two practical demands, current solutions can be categorized into two sequential manners: ``Compress before Enhance~(CbE)'' or ``Enhance before Compress~(EbC)''. 
However, both of them are not suitable since: (1) Error accumulation in the individual models plagues sequential solutions. Especially, once low-light images are compressed by existing general lossy image compression approaches, useful information~(\textit{e.g.}, texture details) would be lost resulting in a dramatic performance decrease in low-light image enhancement. (2) Due to the intermediate process, the sequential solution introduces an additional burden resulting in low efficiency.
We propose a novel joint solution to simultaneously achieve a high compression rate and good enhancement performance for low-light images with much lower computational cost and fewer model parameters. 
We design an end-to-end trainable architecture, which includes the main enhancement branch and the signal-to-noise ratio~(SNR) aware branch. 
Experimental results show that our proposed joint solution achieves a significant improvement over different combinations of existing state-of-the-art sequential ``Compress before Enhance'' or ``Enhance before Compress'' solutions for low-light images, which would make lossy low-light image compression more meaningful.
The project is publicly available at: \url{https://github.com/CaiShilv/Joint-IC-LL}.
\end{abstract}

\begin{figure*}[t]
\centering
\includegraphics[width=\linewidth, height=4.5cm]{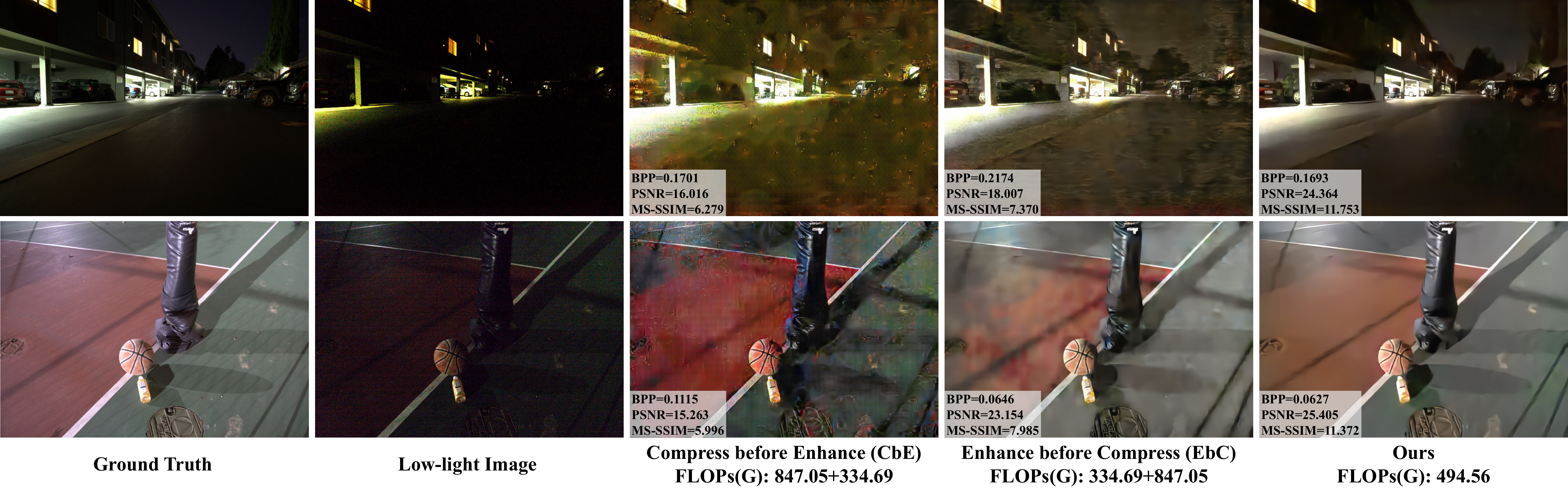}
\caption{Compared with sequential solutions~(``Compress before Enhance~(CbE)'' and ``Enhance before Compress~(EbC)''), our proposed joint solution has significantly greater advantages in terms of PSNR, MS-SSIM, and computational cost with even lower bits per pixel~(BPP). 
As shown, our joint solution makes lossy low-light image compression meaningful with much better visibility for visual perception.
In this teaser figure, the compression and low-light enhancement methods of sequential solutions are Cheng~\cite{cheng2020learned} and Xu2022~\cite{xu2022snr} respectively. 
The example images in the figure are from the SID dataset~\cite{chen2018learning}. For more comparison qualitative results, please refer to the supplementary material.}
\label{head_figure}
\end{figure*}

\section{Introduction}
Low-light images are prevalent in the real world since they are inevitably captured under sub-optimal conditions~(\textit{e.g.}, back, uneven, or dim lighting) or technical limitations~(\textit{e.g.}, limited exposure time).
Low-light images present challenges for human perception and subsequent downstream vision tasks due to unsatisfied visibility.
Therefore, low-light image enhancement is usually employed.
In recent years, the success of learning-based low-light image enhancement~\cite{lore2017llnet, xu2022snr, ma2022toward} has been compelling thus attracting growing attention.
\par
In practical applications, lossy image compression is also crucial for media storage and transmission. Many traditional standards~(\textit{e.g.}, JPEG~\cite{wallace1992jpeg}, JPEG2000~\cite{rabbani2002jpeg2000}, BPG~\cite{bpg}, and Versatile Video Coding~(VVC)~\cite{vvc}) have been proposed and widely used. In recent years, learning-based lossy image compression methods~\cite{cheng2020learned, he2022elic, xie2021enhanced, wang2022neural, liu2023learned} have developed rapidly and outperformed traditional standards in terms of performance metrics, such as the peak signal-to-noise ratio~(PSNR) and the multi-scale structural similarity index~(MS-SSIM).
\par
Whereas, lossy low-light image compression is required in many actual systems as well~(\textit{e.g.}, nighttime autonomous driving and visual surveillance), while little research has been conducted in the academic community on this practical topic.
Current engineering solutions can be categorized into two manners: ``Compress before Enhance~(CbE)'' and ``Enhance before Compress~(EbC)''.
However, existing sequential solutions have at least two major drawbacks: (1) Error accumulation and loss of information in the individual models plague sequential solutions~(see Figure~\ref{head_figure}). In particular, the loss of useful detail information in low-light images after compression severely degrades enhancement performance. Off-the-shelf lossy image compression methods often lack adaptability to low-light images. (2) Sequential solutions introduce additional computational costs due to intermediate results, resulting in low efficiency. Therefore, in this work, we try to answer an important question: \textbf{Can we construct a joint solution of low-light image compression and enhancement, which would achieve high visual quality of reconstructed image under both low computational cost and bits per pixel~(BPP)? Or simply say, can we make lossy low-light image compression more meaningful?}
\par
Based on these considerations, in this work, we propose a novel joint solution for low-light image compression and enhancement. 
We design an end-to-end trainable two-branch architecture with the main enhancement branch for obtaining compressed domain features and the signal-to-noise ratio~(SNR) aware branch for obtaining local/non-local features. Then, the local/non-local features are fused with the compressed domain features to generate the enhanced features for jointly compressing and enhancing low-light images simultaneously.
Finally, the enhanced image is reconstructed by the main decoder.
Our proposed joint solution achieves significant advantages compared to sequential ones, please see Figure~\ref{head_figure} for visualization. More comparison results are included in the supplementary material.
In summary, the contributions of this work are as follows:
\begin{itemize}
\item A joint solution of low-light image compression and enhancement is proposed with much lower computational cost compared to sequential ones.
\item Thanks to the end-to-end trainable two-branch architecture, the joint solution has the ability to achieve high visual quality of reconstructed images with low BPP.
\item Since there is no off-the-shelf joint solution, we compare our model with sequential CbE and EbC solutions~(different combinations and orders of three compression and two enhancement methods respectively) on four datasets to verify the superiority of our joint solution.
\end{itemize}

\section{Related Works}
\paragraph{Learning-based lossy image compression.}
Learning-based image compression methods have shown great potential, which has led to a growing interest among researchers in this field.
Lossy image compression usually contains transform, quantization, and entropy coding. These three components have been studied by many researchers.
\par
There are some works that focus on quantization. Works~\cite{balle2016end, balle2018variational} used the additive uniform noise $\mathcal{U}(-0.5, 0.5)$ instead of the actual quantization during the training. 
Agustsson \textit{et al.}~\cite{agustsson2017soft} proposed soft-to-hard vector quantization to replace scalar quantization. 
Dumas \textit{et al.}~\cite{dumas2018autoencoder} aimed to learn the quantization step size for each latent feature map. 
Zhang and Wu~\cite{zhang2023lvqac} proposed a Lattice Vector Quantization scheme coupled with a spatially Adaptive Companding~(LVQAC) mapping.
\par
Some works focus on the transform, \textit{e.g.}, generalized divisive normalization~(GDN)~\cite{ balle2015density, Balle2016pcs, balle2016end}, residual block~\cite{theis2017lossy}, attention module~\cite{cheng2020learned, zhou2019end},  non-local attention module~\cite{chen2021end}, attentional multi-scale back projection~\cite{gao2021neural},  window attention module~\cite{zou2022devil}, stereo attention module~\cite{wodlinger2022sasic}, and expanded adaptive scaling normalization~(EASN)~\cite{shin2022expanded} have been used to improve the nonlinear transform. 
Invertible neural network-based architecture~\cite{cai2022high, helminger2020lossy, ho2021anfic, ma2019iwave, ma2022end, xie2021enhanced} and transformer-based architecture~\cite{qian2022entroformer, zhu2021transformer, zou2022devil, liu2023learned} also have been utilized to enhance the modeling capacity of the transforms.
\par
Some other works aim to improve the efficiency of entropy coding, \textit{e.g.}, scale hyperprior entropy model~\cite{balle2018variational}, channel-wise entropy model~\cite{minnen2020channel}, context model~\cite{lee2018context, mentzer2018conditional, minnen2018joint}, 3D-context model~\cite{guo20203}, multi-scale hyperprior entropy model~\cite{hu2021learning}, discretized Gaussian mixture model~\cite{cheng2020learned}, checkerboard context model~\cite{he2021checkerboard}, split hierarchical variational compression~(SHVC)~\cite{ryder2022split}, information transformer~(Informer) entropy model~\cite{kim2022joint}, bi-directional conditional entropy model~\cite{lei2022deep}, unevenly grouped space-channel context model~(ELIC)~\cite{he2022elic}, neural data-dependent transform~\cite{wang2022neural}, multi-level cross-channel entropy model~\cite{guo2022practical}, and multivariate Gaussian mixture model~\cite{zhu2022unified}. 
By constructing more accurate entropy models, these methods have achieved greater compression efficiency.
\par
However, existing learning-based compression methods typically do not consider the impact on images of low-light conditions in their design. They may cause unsatisfied image quality and subsequent visual perception problems after decompression due to the loss of detailed information.

\paragraph{Learning-based low-light image enhancement.}
Many learning-based low-light image enhancement methods~\cite{cai2018learning, guo2020zero, jiang2021enlightengan, jin2022unsupervised, kim2021representative, liu2021retinex, lore2017llnet, ma2022toward, ren2019low, wang2021real, wang2022low, wu2022uretinex, xu2022snr,  xu2020learning, yan2014learning, yan2016automatic, yang2021band, yang2021sparse, zamir2020learning, zeng2020learning, zhang2021beyond, zhang2022deep, zhao2021deep, zheng2021adaptive} have been proposed with compelling success in recent years.
\par
For supervised methods, Zhu \textit{et al.}~\cite{zhu2020eemefn} proposed a two-stage method called EEMEFN, which comprised muti-exposure fusion and edge enhancement. Xu \textit{et al.}~\cite{xu2020learning} proposed a frequency-based decomposition-and-enhancement model network. It first learned to recover image contents in a low-frequency layer and then enhanced high-frequency details according to recovered contents. Sean \textit{et al.}~\cite{moran2020deeplpf} introduced three different types of deep local parametric filters to enhance low-light images. 
\par
For semi-supervised methods, Yang \textit{et al.}~\cite{yang2020fidelity} proposed the semi-supervised deep recursive band network~(DRBN) to extract a series of coarse-to-fine band representations of low-light images. The DRBN was extended by using Long Short Term Memory~(LSTM) networks and obtaining better performance~\cite{yang2021band}. 
\par
For unsupervised methods, Jiang \textit{et al.}~\cite{jiang2021enlightengan} proposed an unsupervised generative adversarial network which was the first work that successfully attempted to introduce unpaired training for low-light image enhancement. Ma \textit{et al.}~\cite{ma2022toward} developed a self-calibrated illumination learning method and defined the unsupervised training loss to improve the generalization ability of the model. Fu \textit{et al.}~\cite{fu2023learning} proposed PairLIE which learned adaptive priors from low-light image pairs.
\par
However, these low-light image enhancement methods currently overlook the mutual influence with image compression, resulting in significant performance degradation once CbE or EbC is conducted~(see Figure~\ref{head_figure}). 
In addition, most low-light image enhancement networks have complex architecture designs, and their architectures are not suited to combine with image compression directly in a joint manner.

\paragraph{Joint solutions.}
It is worth noting that, in some other image processing tasks, joint solutions have been verified as an effective alternative to sequential ones with promising results. These joint solutions alleviate the error accumulation effect in the pipeline process.
The success of the joint solution of multiple tasks using a single network architecture has attracted the attention of researchers in the development of deep learning.
There are some works studied for joint solutions have made progress including joint denoising and demosaicing~\cite{ehret2019joint, gharbi2016deep}, joint image demosaicing, denoising and super-resolution~\cite{xing2021end},  joint low-light enhancement and denoising~\cite{lu2022progressive}, and joint low-light enhancement and deblurring~\cite{zhou2022lednet}. 
Recently, some works~\cite{cheng2022optimizing, de2022satellite, ranjbar2022joint} optimize image processing and image compression jointly. Cheng \textit{et al.}~\cite{cheng2022optimizing} jointed image compression and denoising to resolve the bits misallocation problem. Jeong \textit{et al.}~\cite{jeong2022rawtobit} proposed the RAWtoBit network~(RBN), which jointly optimizes camera image signal processing and image compression. Qi \textit{et al.}~\cite{qi2023real} proposed a framework for real-time 6K rate-distortion-aware image rescaling which could reconstruct a high-fidelity HR image from the JPEG thumbnail.
\par
Nevertheless, the aforementioned methods are ill-suited for low-light image compression and enhancement. This interesting issue has received limited research attention within the academic community yet.

\begin{figure*}[t]
\centering
\includegraphics[width=\linewidth,height=5.5cm]{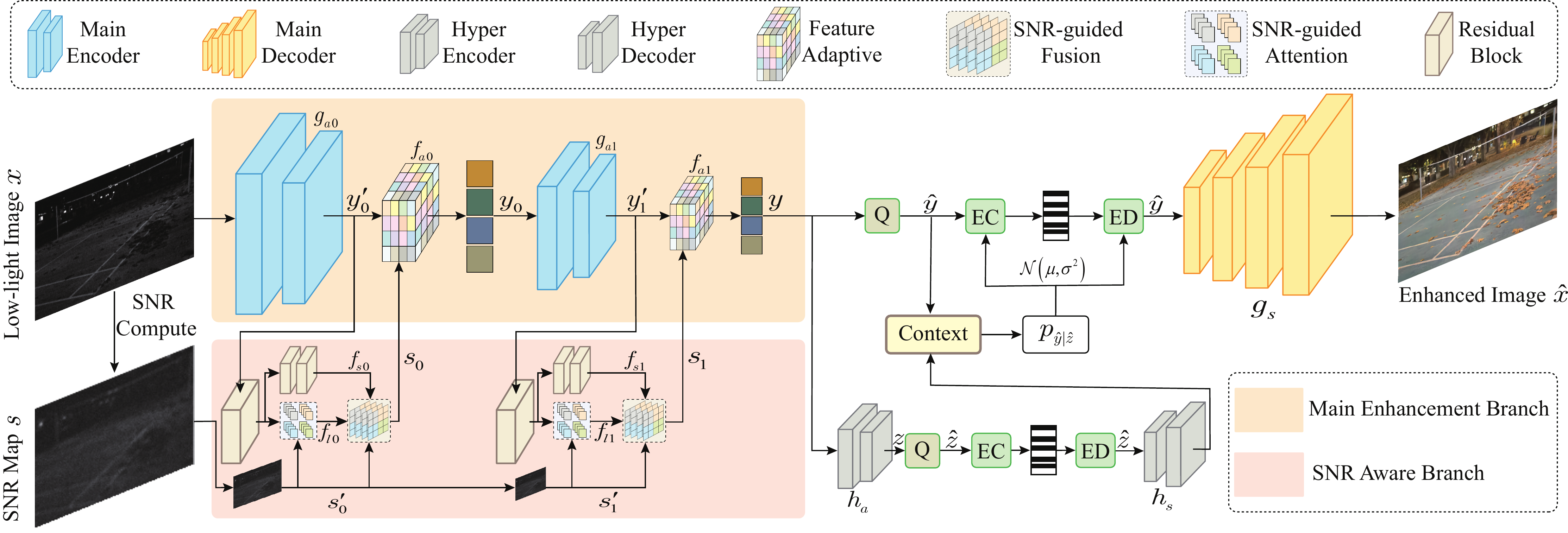}
\caption{The network architecture of our joint solution of low-light image compression and enhancement. The left half of the figure contains two branches, the ``Main Enhancement Branch" and the ``SNR Aware Branch". The low-light image is fed into the ``Main Enhancement Branch" to obtain the two-level enhanced compressed domain features ($y_{0}$/$y$) via ``Feature Adaptive" modules~($f_{a0}$/$f_{a1}$). The ``SNR Aware Branch" obtains local/non-local information by the SNR-map $s$ and compressed domain features~($y_0^{\prime}$/$y_{1}^{\prime}$). The right half of the figure contains the main decoder, entropy models, context model, and hyper encoder/decoder commonly used in recent learning-based compression methods~\cite{minnen2018joint, cheng2020learned}. ``/'' means ``or'' in this paper.}
\label{network_architecture}
\end{figure*}

\section{Methodology}
\subsection{Problem Formulation}
\paragraph{Lossy image compression.}
We briefly introduce the formulation of the learning-based lossy image compression first.
In the widely used variational auto-encoder based framework~\cite{balle2018variational}, the source image $x$ is transformed to the latent representation $y$ by the parametric encoder $g_{a}(x; \phi_{a})$. The latent representation $y$ is quantized to discrete value $\hat{y}$ which is losslessly encoded to bitstream using entropy coders~\cite{duda2013asymmetric, witten1987arithmetic}. During the decoding, $\hat{y}$ is obtained through entropy decoding the bitstream. Finally, $\hat{y}$ is inversely transformed to the reconstructed image $\hat{x}$ through the parametric decoder $g_{s}(\hat{y}; \phi_{s})$. In fact, the optimization of the image compression model for the rate-distortion performance can be realized by minimizing the expectation Kullback-Leibler~(KL) divergence between intractable true posterior $p_{\hat{y}|x}(\hat{y}|x)$ and parametric variational density $q(\hat{y}|x)$ over the data distribution $p_{x}$~\cite{balle2018variational}:
\begin{small}
\begin{equation}
\centering
\begin{aligned}
\mathbb{E}_{x\sim p_{x}}D_{KL}[q(\hat{y}|x)||p({\hat{y}|x})]=\mathbb{E}_{x\sim       p_{x}}\mathbb{E}_{\hat{y}\sim q}\bigl[\log q(\hat{y}|x) \\ 
-\underbrace{\log p_{x|\hat{y}}(x|\hat{y})}_{\text{weighted distortion}} - \underbrace{\log p_{\hat{y}}(\hat{y})}_{\text{rate}}\bigr] + \textrm{const},
\label{KL_eq}
\end{aligned}
\end{equation}
\end{small}
where $D_{KL}[\cdot ||\cdot ]$ is the KL divergence. Given the transform parameter $\phi_{a}$, the transform $y=g_{a}(x; \phi_{a})$~(from $x$ to $y$) is determined and the process of quantizing $y$ is equivalent to adding uniform distribution $\mathcal{U}(-1/2, 1/2)$ for relaxation. Therefore, $q({\hat{y}|x})=\prod_{i} \mathcal{U}(y_{i}-1/2, y_{i}+1/2)$ and the first term $\log q(\hat{y}|x)=0$. The second term $\log p_{x|\hat{y}}(x|\hat{y})$ is the expected distortion between source image $x$ and reconstructed image $x^{\circ}$. The third term reflects the cost of entropy encoding discrete value $\hat{y}$.
\par
In order to make the second term of Eq.~\ref{KL_eq} easier to calculate. Suppose that the likelihood is give by $p(x|\hat{y})= \mathcal{N}(x|x^{\circ}, (\mathfrak{p} \cdot \lambda)^{-1}1)$. In addition, considering the introduction of scale hyperprior. Similar to previous works~\cite{balle2018variational, cheng2020learned}, the rate-distortion objective function can be written as:
\begin{small}
	\begin{equation}
		\mathbb{E}_{x\sim p_{x}}\mathbb{E}_{\hat{y},\hat{z}\sim q}\bigl[ \lambda \cdot \left\|x - x^{\circ}\right\|_{\mathfrak{p}}^{\mathfrak{p}} - \log p_{\hat{y}|\hat{z}}(\hat{y}|\hat{z}) - \log p_{\hat{z}}(\hat{z}) \bigr],
		\label{rate_distortion_equation}
	\end{equation}
\end{small}
where the parameter $\lambda$ is the trade-off between distortion and compression levels. If the value of $\mathfrak{p}=2$, the first term is the mean square error~(MSE) distortion. The additional side information $\hat{z}$ is used to capture spatial dependencies.
\paragraph{\mbox{Supervised learning-based low-light image enhancement.}}
The low-light image refer as $x\in \mathbb{R}^{3\times h\times w}$. $h$ and $w$ denote the height and width of the low-light image respectively. The low-light enhancement processing can be expressed as:
\begin{equation}
	\bar{x}=\mathcal{G}(x; \theta),
\end{equation}
where the $\bar{x}$ denotes the reconstructed low-light enhancement image. $\theta$ represents the learnable parameters of the neural network $\mathcal{G}$. The optimization of the learning-based low-light image enhancement model is done by minimizing loss to learn the optimal network parameters $\hat{\theta}$:
\begin{equation}
\hat{\theta}=\mathop{\textrm{argmin}}_{\theta} \mathcal{L}_{e}(\mathcal{G}(x; \theta), x^{gt})=\mathop{\textrm{argmin}}_{\theta} \mathcal{L}_{e}(\bar{x}, x^{gt}).
\label{loss_enhance}
\end{equation}
The loss function $\mathcal{L}_{e}(\cdot,\cdot)$ usually can use $L_{1}$, $L_{2}$, or Charbonnier~\cite{lai2018fast} loss, etc. The network parameters $\theta$ can be optimized by minimizing the error between the reconstructed image $\bar{x}$ and the ground truth image $x^{gt}$.

\paragraph{Joint formulation.}
Based on Eq.~\ref{rate_distortion_equation} and Eq.~\ref{loss_enhance}, we further develop the joint formulation of image compression and low-light image enhancement by simultaneously optimizing the rate distortion and the similarities between enhanced and ground truth images as follows:
\begin{small}
\begin{equation}
\begin{aligned}
	 \mathcal{L}=&\lambda_{d} \cdot \mathcal{D}(x^{gt},\hat{x}) + \mathcal{R}(\hat{y}) + \mathcal{R}(\hat{z}) \\
	 =&\lambda_{d} \cdot \mathbb{E}_{x\sim p_{x}}\bigl[\left\|x^{gt} - \hat{x}\right\|_{\mathfrak{p}}^{\mathfrak{p}}\bigr]\\ 
	 & - \mathbb{E}_{\hat{y} \sim q_{\hat{y}}}\bigl[\log p_{\hat{y}|\hat{z}}(\hat{y}|\hat{z})\bigr] - \mathbb{E}_{\hat{z} \sim q_{\hat{z}}}\bigl[\log p_{\hat{z}}(\hat{z})\bigr].
\label{Joint_loss1}
\end{aligned}
\end{equation}
\end{small}
The first term $\mathcal{D}(x^{gt},\hat{x})$ measures distortion between the ground truth image $x^{gt}$ and the enhanced image $\hat{x}$. The second term $\mathcal{R}(\hat{y})$ and third term $ \mathcal{R}(\hat{z})$ denote the compression levels. $\lambda_{d}$ denotes the weighting coefficient, which is the trade-off between compression levels and distortion. If $\mathfrak{p} = 2$, the first term is mean square error~(MSE) distortion.

\subsection{Framework}
\paragraph{Overall workflow.}
Figure~\ref{network_architecture} shows an overview of the network architecture of our proposed joint solution of low-light image compression and enhancement. The low-light image $x$ is transformed to the enhanced compressed domain features $y$ by main encoders $g_{a0}$ and $g_{a1}$ with SNR-guided feature adaptive operations. Then $y$ is quantized to the discrete enhanced compressed domain features $ \hat{y}$ by the quantizer Q. The uniform noise $\mathcal{U}(-1/2, 1/2)$ is added to the enhanced compressed domain features $y$ instead of non-differentiable quantization operation during the training and rounding the enhanced compressed domain features $y$ during testing~\cite{balle2018variational}.
\par
We use the hyper-prior scale~\cite{balle2018variational, minnen2018joint} module to effectively estimate the distribution $p_{\hat{y}|\hat{z}} \sim \mathcal{N}(\mu, \sigma^2)$ of the discrete enhanced compressed domain features $\hat{y}$ by generating parameters~($\mu$ and $\sigma$) of the Gaussian entropy model to support entropy coding/decoding~(EC/ED). The latent representation $z$ is quantized to $\hat{z}$ by the same quantization strategy as the enhanced features $y$. The distribution of discrete latent representation $\hat{z}$ is estimated by the factorized entropy model~\cite{balle2016end}. The range asymmetric numeral system~\cite{duda2013asymmetric} is used to losslessly compress discrete enhanced features $\hat{y}$ and latent representation $\hat{z}$ into bitstreams. The decoded enhanced features $\hat{y}$ obtained by the entropy decoding are fed into the main decoder $g_{s}$ to reconstruct the enhanced image $\hat{x}$. It is worth noting that the proposed joint solution integrates compression and low-light enhancement into a single process that performs both tasks simultaneously, achieving excellent performance while significantly reducing the computational cost.
\paragraph{Two branch architecture.}
Our proposed joint solution includes two branches. The first branch is the signal-to-noise ratio~(SNR) aware branch. The SNR map $s$ is achieved by employing a no-learning-based denoising operation~(refer Eq.~\ref{SNR_map_calculate}) which is simple yet effective. Local/non-local information on the low-light image is obtained through the SNR-aware branch. 
The second branch is the main enhancement branch, the compressed domain features~($y_{0}^{\prime}$/$y_{1}^{\prime}$) combine with the local/non-local information~($s_{0}$/$s_{1}$) generated by the SNR-aware branch to obtain the enhanced compressed domain features~($y_{0}$/$y$).

\subsection{Enhanced Compressed Domain Features}
As Figure~\ref{network_architecture} shows, the SNR map $s \in \mathbb{R}^{h \times w}$ is estimated from the low-light image $x\in \mathbb{R}^{3\times h \times w}$. The calculation process starts by converting low-light image $x$ into grayscale image $\dot{x} \in \mathbb{R}^{h \times w}$ and then proceeds as follows:
\begin{equation}
	\ddot{x} = kernel(\dot{x}), \quad n=abs(\dot{x} - \ddot{x}), \quad s = \frac{\ddot{x}}{n},
	\label{SNR_map_calculate}
\end{equation}
where $kernel(\cdot)$ denotes averaging local pixel groups operation, $abs(\cdot)$ denotes taking absolute value function.
\par
The SNR map $s$ is processed by the residual block module~(``Residual Block" in Figure~\ref{network_architecture}) and transformer-based module~(``SNR-guided Attention" in Figure~\ref{network_architecture}) with generating the local features~($f_{s0}$/$f_{s1}$) and the non-local features~($f_{l0}$/$f_{l1}$) inspired by the work~\cite{xu2022snr}. Local and non-local features are fused. It is illustrated in ``SNR-guided Fusion" of Figure~\ref{network_architecture} and is calculated as follows:
 \begin{equation}
 \begin{split}
 	&s_{0}=f_{s0} \times s_{0}^{\prime} + f_{l0} \times (1 - s_{0}^{\prime}),\\
 	&s_{1}=f_{s1} \times s_{1}^{\prime} + f_{l1} \times (1 - s_{1}^{\prime}),
 \end{split}
 \end{equation}
 where $s_{0}^{\prime}$ and $s_{1}^{\prime}$ are resized from SNR map $s$ according to the shape of corresponding features~($f_{s0}$/$f_{s1}$/$ f_{l0}$/$f_{l1}$). $s_{0}$ and $s_{1}$ are SNR-aware fusion features.
 \par
\begin{figure}[t]
	\centering
	\includegraphics[width=0.85\linewidth]{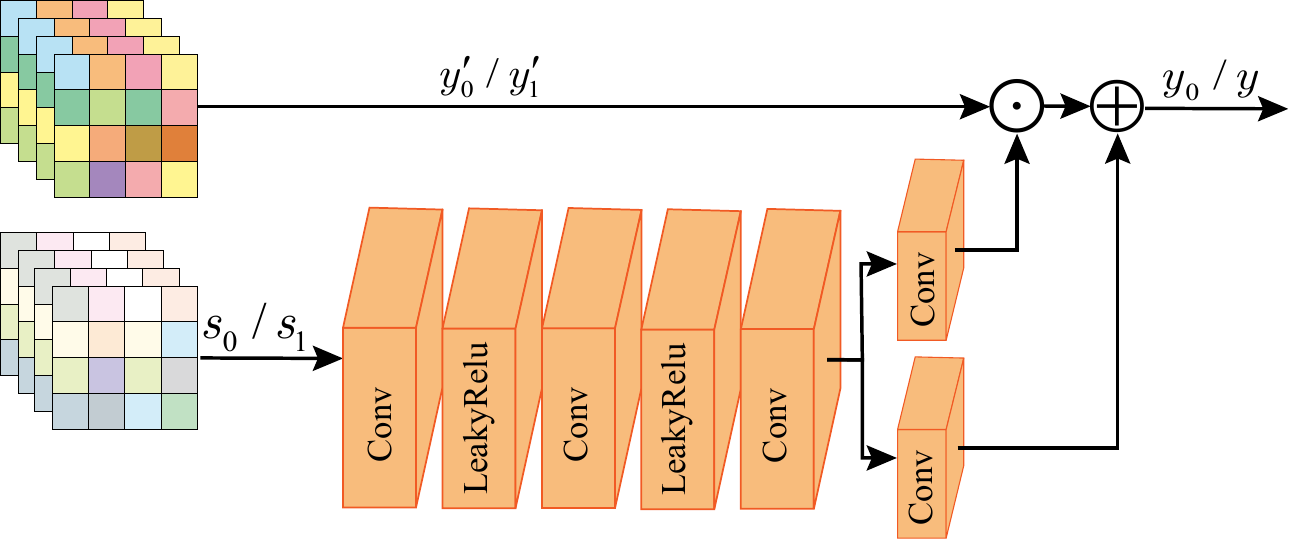}
	\caption{Architecture details of the ``Feature Adaptive" module. SNR-aware fusion features~($s_{0}$/$s_{1}$) act as a condition on the compressed domain features~($y_0^{\prime}$/$y_{1}^{\prime}$) to generate enhanced features~($y_{0}$/$y$). $\odot$ denotes the Hadamard product and $\oplus$ denotes the addition by element.}
	\label{feature_adaptive}
\end{figure}

 Since the SNR map $s$ is unavailable in the decoding process, we consider enhancing the features $y_{0}$ and $y$ in the compressed domain instead of the manner~\cite{xu2022snr} using the decoded domain. 
 Thus, the enhanced image $\hat{x}$ can be obtained by decoding the enhanced features $\hat{y}$ directly. The compressed domain features~($y_0^{\prime}$/$y_{1}^{\prime}$) are enhanced by ``Feature Adaptive" modules~(refered as $f_{a0}$/$f_{a1}$), shown in Figure~\ref{network_architecture}, and their details are shown in Figure~\ref{feature_adaptive}.

\subsection{Training Strategy}
In our experiments, we observe that training both image compression and low-light image enhancement tasks jointly at the beginning results in convergence problems. Thus, we adopt the two-stage training.
\paragraph{Pre-train without SNR-aware branch.}
We pre-train the model without joining the signal-to-noise ratio~(SNR) aware branch. In this case, the network architecture is similar to the Cheng2020-anchor~\cite{cheng2020learned} of the CompressAI library~\cite{begaint2020compressai} implementation. The rate-distortion loss is:
\begin{small}
\begin{equation}
\begin{aligned}
	 \mathcal{L}=&\lambda_{d} \cdot \mathcal{D}(x,\hat{x}) + \mathcal{R}(\hat{y}) + \mathcal{R}(\hat{z}) \\
	 =&\lambda_{d} \cdot \mathbb{E}_{x\sim p_{x}}\bigl[\left\|x - \hat{x}\right\|_{\mathfrak{p}}^{\mathfrak{p}}\bigr]\\ 
	 & - \mathbb{E}_{\hat{y} \sim q_{\hat{y}}}\bigl[\log p_{\hat{y}|\hat{z}}(\hat{y}|\hat{z})\bigr] - \mathbb{E}_{\hat{z} \sim q_{\hat{z}}}\bigl[\log p_{\hat{z}}(\hat{z})\bigr],
\label{Joint_loss2}
\end{aligned}
\end{equation}
\end{small}
where $x$ and $\hat{x}$ denote the original image and decoded image respectively. We set the $\lambda_{d}=0.0016$. It is worth noting that the parameter $\mathfrak{p}$ of  the first term $\mathbb{E}_{x\sim p_{x}}\bigl [\left\|x - \hat{x}\right\|_{\mathfrak{p}}^{\mathfrak{p}}\bigr ]$ is equal to 2. That means, the distortion loss $\mathcal{D}(x,\hat{x})$ is the MSE loss.
\paragraph{Train the entire network.}
We train the entire network by loading the pre-trained model parameters. The joint loss function is Eq.~\ref{Joint_loss1}. The parameter $\mathfrak{p}$ of the first term $\mathbb{E}_{x\sim p_{x}}\bigl [\left\|x^{gt} - \hat{x}\right\|_{\mathfrak{p}}^{\mathfrak{p}}\bigr ]$ is equal to 1. That means, we employ $L_{1}$ as the distortion loss $\mathcal{D}(x^{gt},\hat{x})$ instead of the MSE loss to ensure stable training, mitigating the risk of encountering the episodic non-convergence problem.

\begin{figure*}[t]
\centering
\includegraphics[width = 0.98\linewidth, height=7.5cm]{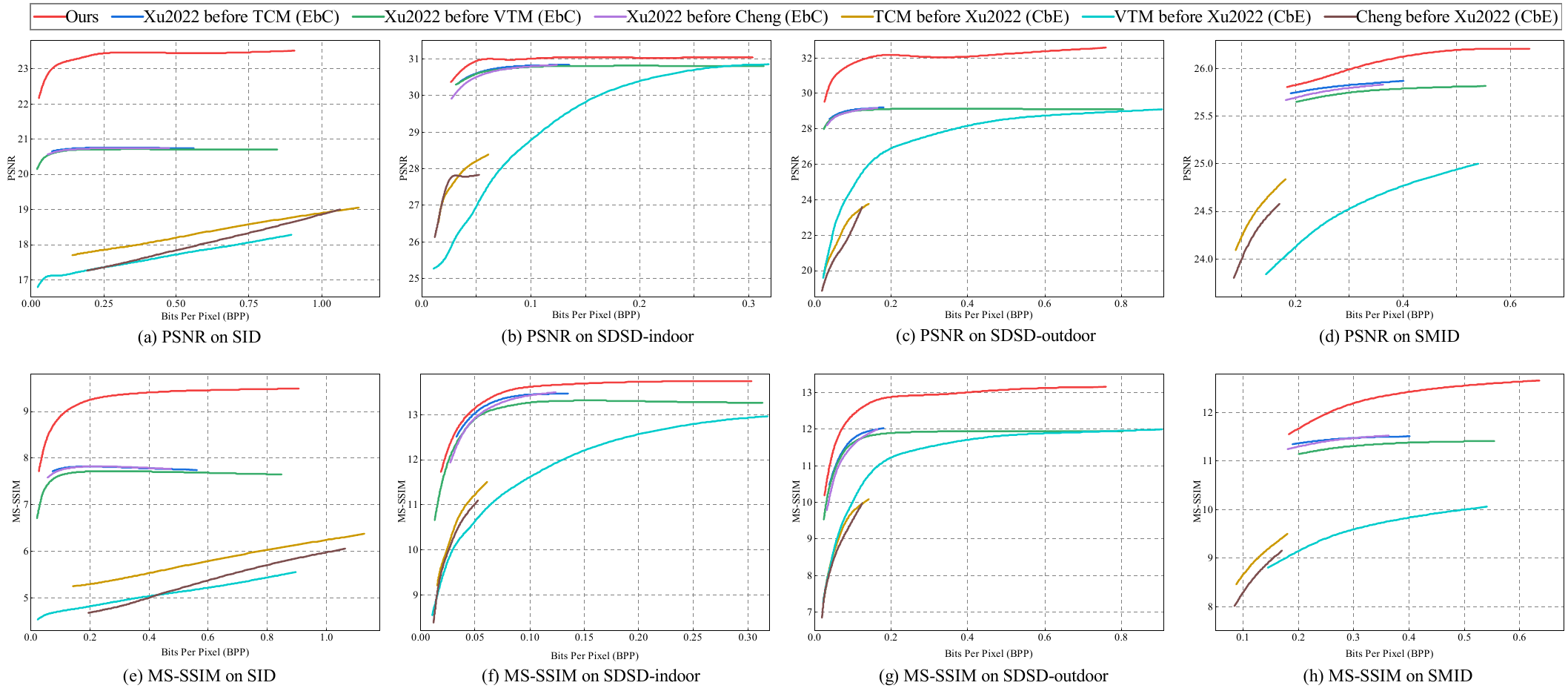}
\caption{Rate-distortion performance curves aggregated over four test datasets. (a)/(b)/(c)/(d) and (e)/(f)/(g)/(h) are results on SID, SDSD-indoor, SDSD-outdoor, and SMID about PSNR and MS-SSIM, respectively. 
Remarkably, we are the first to address the problem of error accumulation and information loss in the joint task of image compression and low-light image enhancement,
so there is no existing method for comparison. We adopt the low-light enhancement method~\cite{xu2022snr} for comparison.
Experimental results obviously show that our proposed joint solution achieves great advantages compared to both ``Compress before Enhance~(CbE)'' and ``Enhance before Compress~(EbC)'' sequential solutions.}
\label{performance_curve}
\end{figure*}

\section{Experiments}
\subsection{Datasets and Implementation Details}
\paragraph{Datasets.}
The Flicker 2W~\cite{liu2020unified} is used in the pre-training and fine-tuning stages for all learning-based methods involved in the comparison. The low-light datasets that we use include SID~\cite{chen2018learning}, SDSD~\cite{wang2021seeing}, and SMID~\cite{chen2019seeing}. The SID and SMID contain pairs of short- and long-exposure images with the resolution of $960 \times 512$. Both SID and SMID have heavy noise because they are captured in extreme darkness. The SDSD~(static version) dataset contains an indoor subset and an outdoor subset with low-light and normal-light pairs. We set up splitting for training and testing based on the previous work~\cite{xu2022snr}. All low-light data are converted to the RGB domain for experiments.

\paragraph{Implementation details.}
We use the image compression anchor model~\cite{cheng2020learned} as our main architecture except for the ``Feature Adaptive" modules and the SNR-aware branch. Randomly cropped patches with a resolution of $512 \times 512$ pixels are used to optimize the model during the pre-training stage. Our implementation relies on Pytorch~\cite{paszke2019pytorch} and the open-source CompressAI PyTorch library~\cite{begaint2020compressai}. 
The networks are optimized using the Adam~\cite{Kingma2015Adam} optimizer with a mini-batch size of 8 for approximately 900000 iterations and trained on RTX 3090 GPUs.
The initial learning rate is set as $10^{\mbox{-}4}$ and decayed by a factor of 0.5 at iterations 500000, 600000, 700000, and 850000. The number of pre-training iteration steps is 150000.
We have a loss cap for each model, so the network will skip optimizing a mini-step if the training loss is above the specified threshold. 
We train our model under 8 qualities, where $\lambda_{d}$ is selected from the set \{0.0001, 0.0002, 0.0004, 0.0008, 0.0016, 0.0028, 0.0064, 0.012\}. To verify the performance of the algorithm, the peak signal-to-noise ratio~(PSNR) and the multi-scale structural similarity index~(MS-SSIM) are used as evaluation metrics. 
We also compare the size of the models and computational cost. 
For better visualization, the MS-SSIM is converted to decibels $(-10log_{10}(1-\textrm{MS-SSIM}))$.

\begin{figure}[t]
\centering
\includegraphics[width = 0.49\linewidth, height= 3.cm]{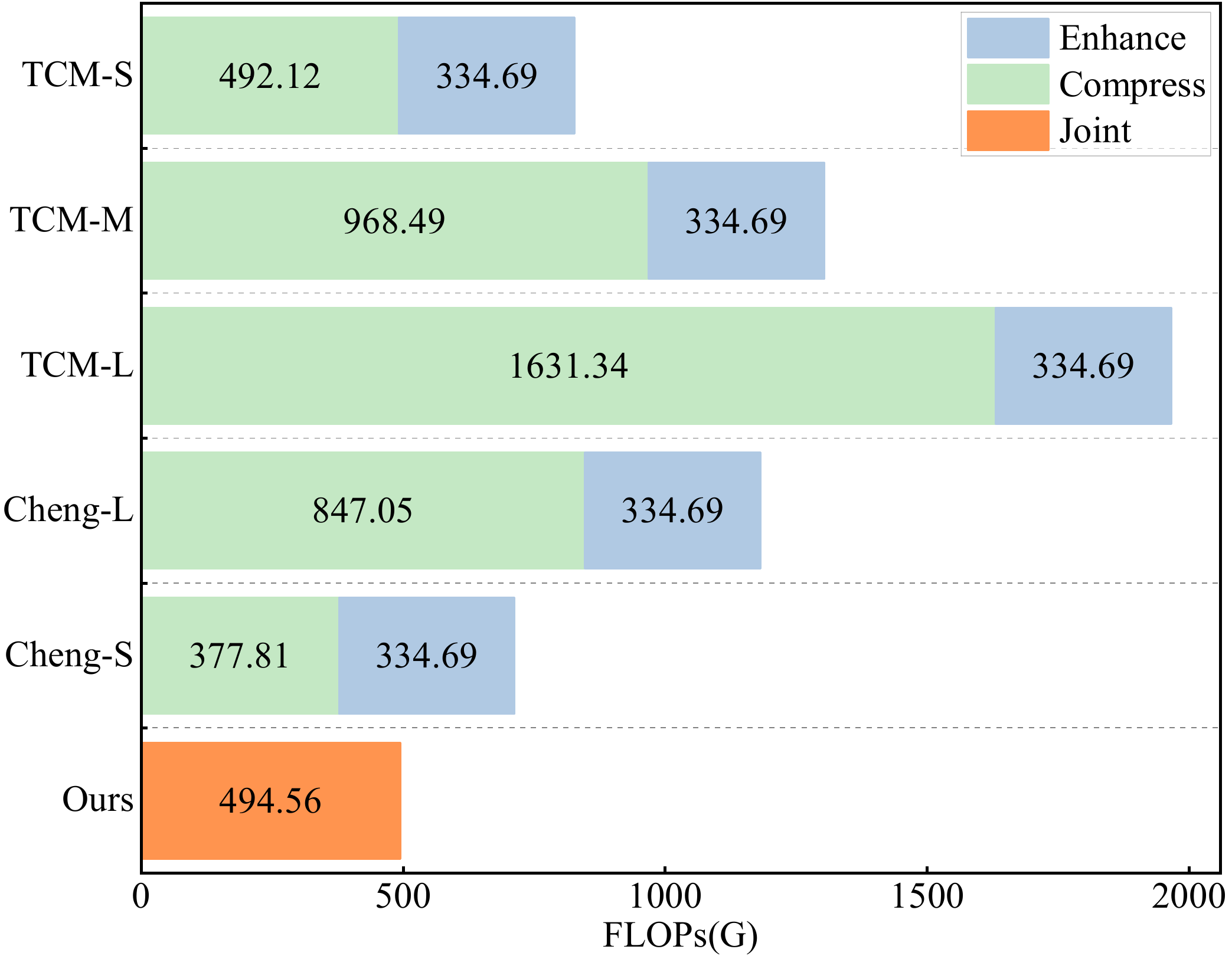}\hfill
\includegraphics[width = 0.49\linewidth, height= 3.cm]{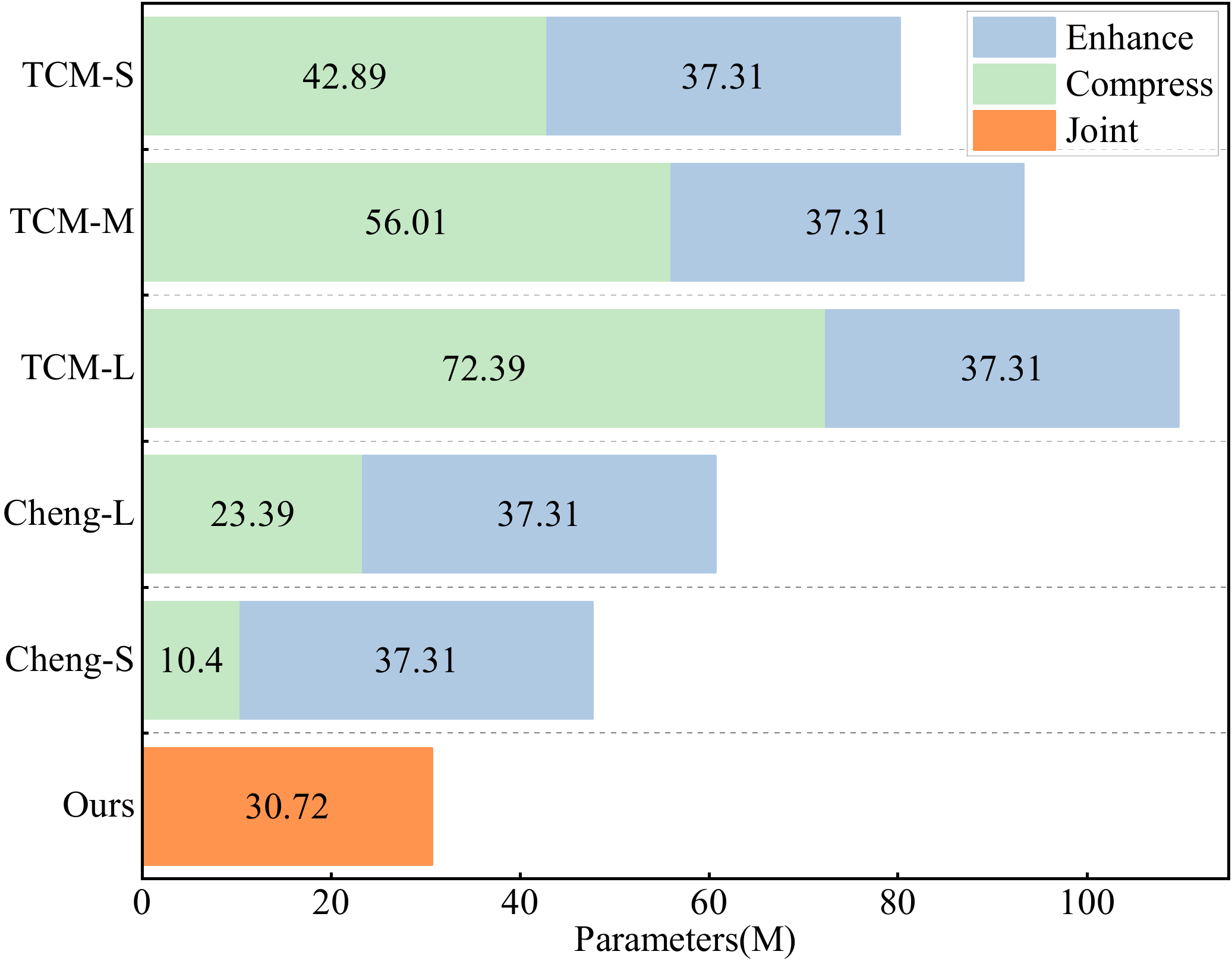}
\caption{Comparison of computational costs and model size. ``TCM-S"/``TCM-M"/``TCM-L" represents the sequential solution of the 64/96/128 channels compression method~\cite{liu2023learned} before the low-light image enhancement method~\cite{xu2022snr}. ``Cheng-S"/``Cheng-L" represents the sequential solution of the 128/192 channels compression method~\cite{cheng2020learned} before the low-light image enhancement method~\cite{xu2022snr}. Obviously, our joint solution has the advantage of lower computational costs and fewer model parameters.}
\label{model_compare}
\end{figure}

\subsection{Algorithm Performance}
\paragraph{Rate-distortion performance.}
Sequential solutions contain individual models of the state-of-the-art low-light enhancement method Xu2022~\cite{xu2022snr}, the state-of-the-art compression method TCM~\cite{liu2023learned}, the typical learning-based compression method Cheng2020-anchor~\cite{cheng2020learned}, and the classical codec method VVC~\cite{vvc}).
The proposed joint solution compares with the six sequential solutions as follows:
(1) ``Xu2022 before TCM~(EbC)'';
(2) ``Xu2022 before VTM~(EbC)'';
(3) ``Xu2022 before Cheng~(EbC)'';
(4) ``TCM before Xu2022~(CbE)'';
(5) ``VTM before Xu2022~(CbE)'';
(6) ``Cheng before Xu2022~(CbE)''.
For brief representation, ``Cheng" denotes the compression method cheng2020Anchor, and ``VTM" denotes the classical codec method VVC.
\par
For image compression methods, we fine-tune the pre-trained Cheng2020-anchor models provided by the CompressAI PyTorch library~\cite{begaint2020compressai} and the models provided by TCM~\cite{liu2023learned} on the Flicker and paired low-light image training datasets for fair comparison. The VCC is implemented by the official Test Model VTM 12.1 with the intra-profile configuration from the official GitHub page to test images, configured with the YUV444 format to maximize compression performance.
For the low-light enhancement method Xu2022, we use the source code obtained from the official GitHub page finetuned on the same paired training datasets for fair comparison.
We show the overall rate-distortion~(RD) performance curves on SID, SDSD-indoor, SDSD-outdoor, and SMID datasets in Figure~\ref{performance_curve}. Our proposed solution~(red curves) achieves great advantages with the common metrics PSNR and MS-SSIM. More qualitative results with quantitative metrics are included in the supplementary material.
\par
Obviously, the error accumulation and loss of information in the individual models plague the sequential solution. Especially, the compressed low-light images with useful information loss make it difficult for the low-light image enhancement method to reconstruct pleasing images.

\paragraph{Computational complexity.}
We compare the computational cost and model size of the proposed joint solution with sequential solutions of the typical learning-based image compression method Cheng2020-anchor~\cite{cheng2020learned}, the state-of-the-art learning-based image compression method TCM~\cite{liu2023learned} and the low-light image enhancement method Xu2022~\cite{xu2022snr}.
As shown in Figure~\ref{model_compare}, the left side of the figure shows the computational cost over an RGB image with the resolution of $960 \times 512$, and the right side of the figure shows the number of model parameters. In our proposed joint solution, the low-light image enhancement and image compression share the same feature extractor/decoder during the encoding/decoding. Thus, the proposed joint solution achieves much lower computational costs and fewer model parameters.

\begin{figure}[t]
\centering
\includegraphics[width = \linewidth]{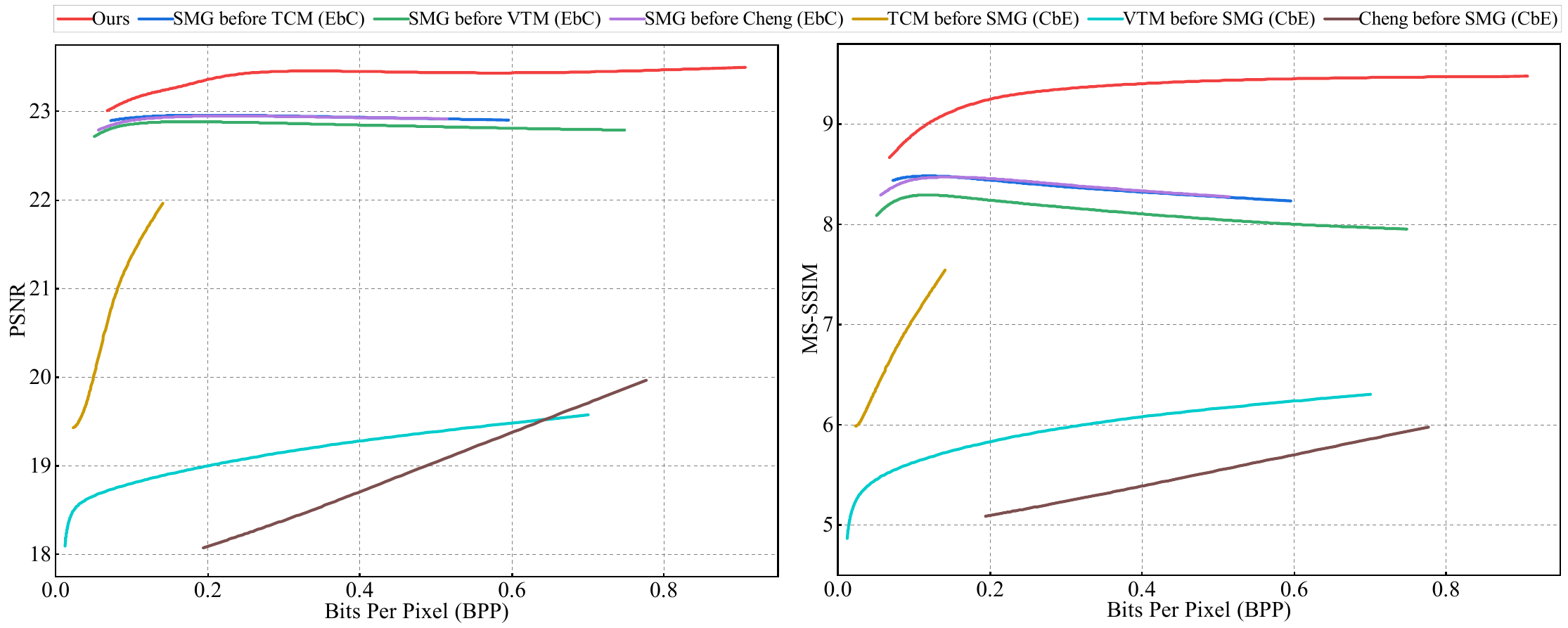}
\caption{We adopt the state-of-the-art low-light enhancement method SMG~\cite{xu2023low} for comparison on the SID dataset.
The results of the experiments show that the proposed joint solution also achieves the greatest advantages compared to the sequential solutions.}
\label{performance_curve_smg}
\end{figure}

\paragraph{Comparison with another enhancement method.}
To further verify the effectiveness of the joint solution, we have also performed comparison experiments with another state-of-the-art low-light image enhancement method SMG~\cite{xu2023low}. 
The proposed joint solution compares with the six sequential solutions as follows: 
(1) ``SMG before TCM (EbC)''; 
(2) ``SMG before VTM (EbC)''; 
(3) ``SMG before Cheng (EbC)''; 
(4) ``TCM before SMG (CbE)''; 
(5) ``VTM before Xu2022 (CbE)''; 
(6) ``Cheng before Xu2022 (CbE)''.
The comparison results on the SID dataset are shown in Figure~\ref{performance_curve_smg}. 
It is worth noting that SMG uses a more complex network structure, implying a higher computational cost. 
The experimental results show that our proposed joint solution consistently has a large advantage over sequential solutions. This indicates that our proposed method can indeed solve the problem of error accumulation and loss of information in sequential solutions.

\begin{figure}[t]
\centering
\includegraphics[width = .49\linewidth, height=3.5cm]{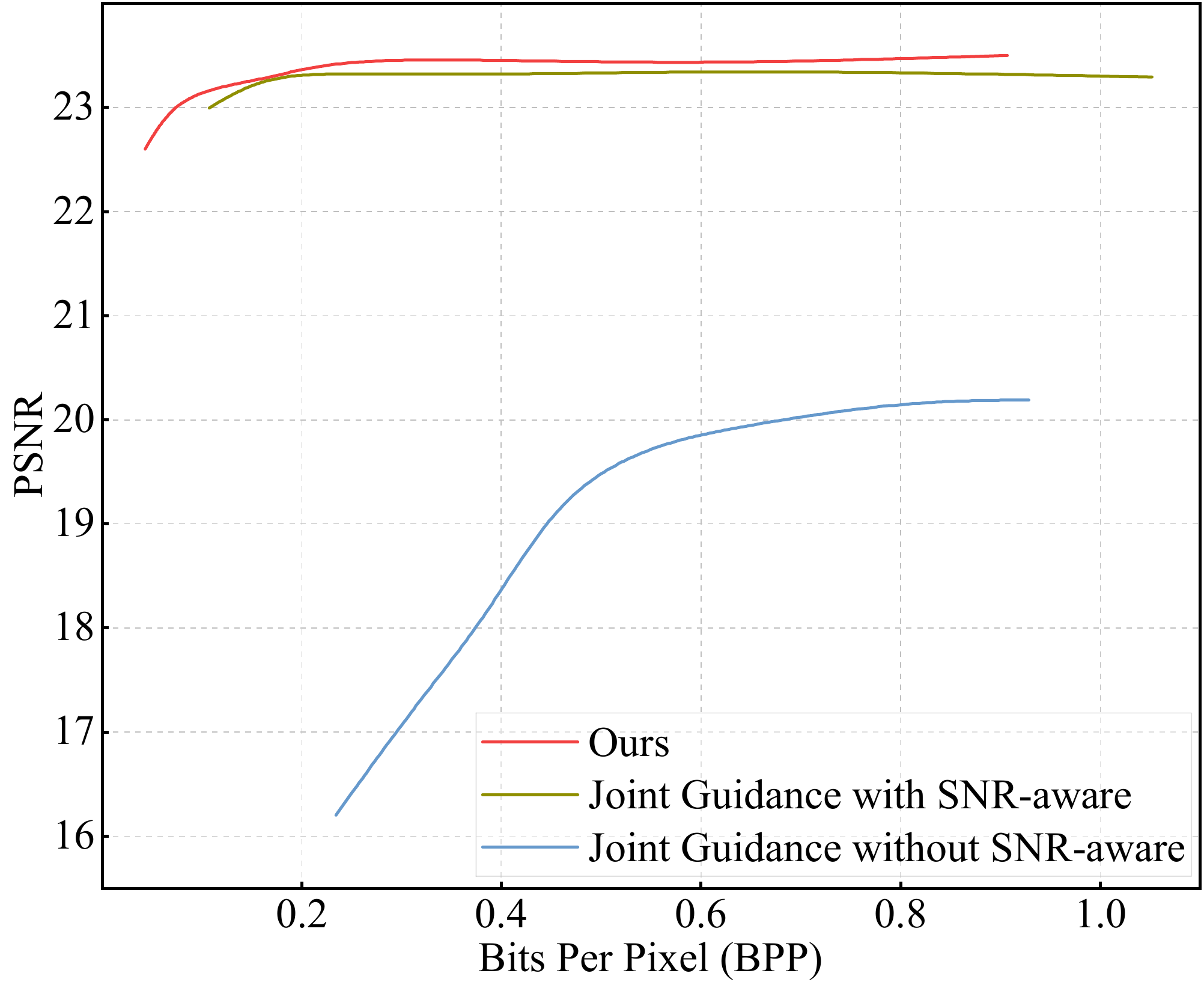}\hfill
\includegraphics[width = .49\linewidth, height=3.5cm]{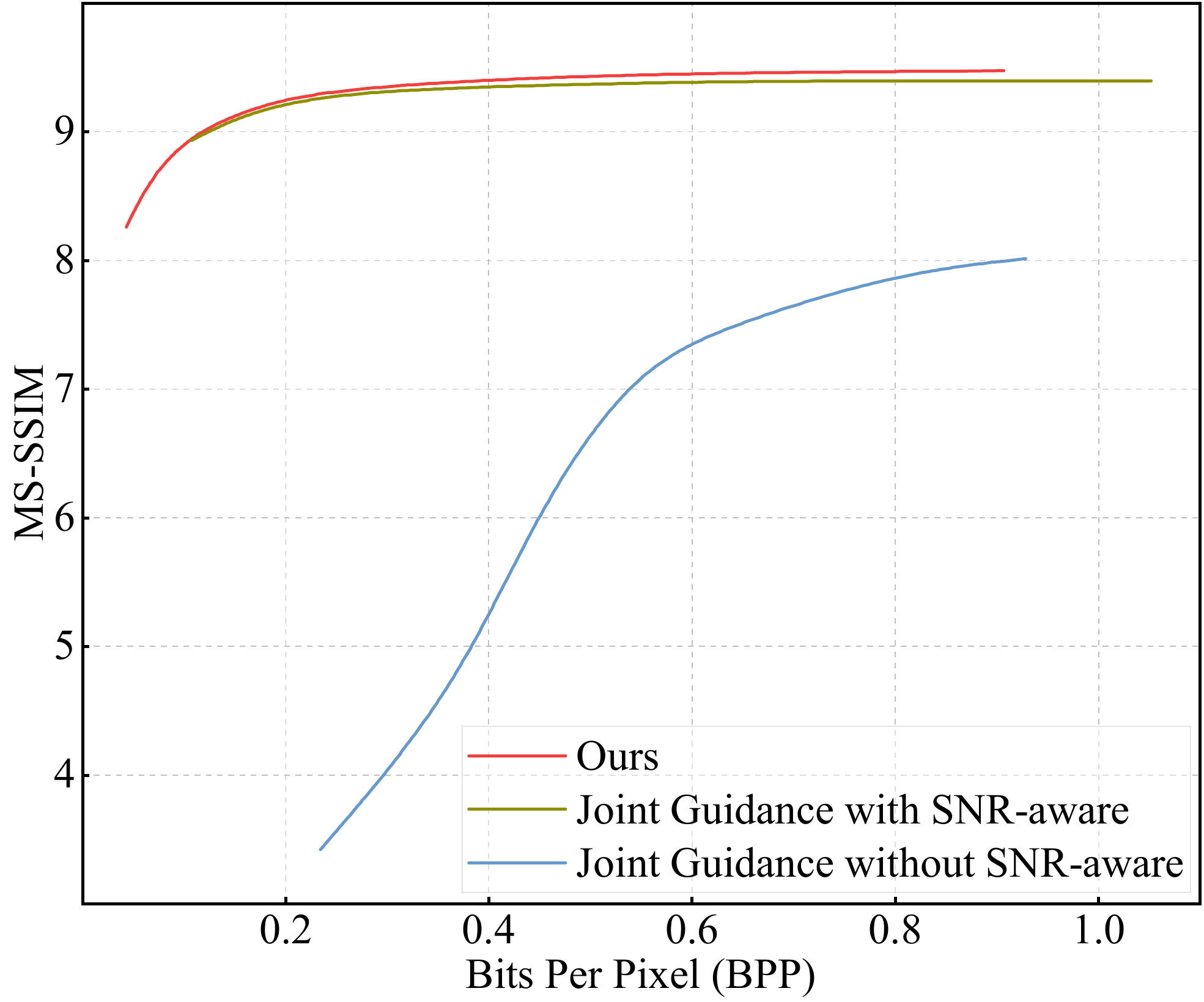}
\caption{The impact of different branches on RD performance. The curves are aggregated on the SID. More experimental results are presented in the supplementary material.}
\label{RD_GW_curve_1}
\end{figure}

\subsection{Analysis}
\paragraph{Impact of the SNR-aware branch.}
The SNR-aware branch can effectively extract local and non-local information from the low-light image by being aware of the signal-to-noise ratio, which is crucial for our low-light image enhancement. To verify the effectiveness of the SNR-aware branch, we remove the SNR-aware branch and add corresponding network modules to the main enhancement branch to achieve low-light image enhancement. We name this method ``Joint Guidance without SNR-aware". The model architecture is similar to DC~\cite{cheng2022optimizing}. More details of this method are given in the supplementary material. 
Figure~\ref{RD_GW_curve_1} shows the results of our method outperforms the ``Joint Guidance without SNR-aware" by a large margin, indicating that the significance and importance of the SNR-aware branch~(red curve vs blue curve).
\paragraph{Joint guidance with SNR-aware.}
To further investigate another training strategy by using the SNR-aware information, we additionally use a three-branch network architecture (named ``Joint Guidance with SNR-aware") for experiments. It has an additional teacher guidance branch during the training stage. Details are shown in the supplementary material. The comparison results are shown in Figure~\ref{RD_GW_curve_1}. The performance of using such a ``Teacher Guidance Branch" is slightly worse than our joint solution~(red curve vs yellow curve), while additionally increasing the computational cost during the training procedure.
That is, our usage of SNR-aware information is more effective and efficient.

\section{Conclusion}
We propose a novel joint solution to make lossy image compression meaningful for low-light images, alleviating the problem of error accumulation when the two tasks are performed in sequential manners. Local and non-local features~(obtained by the SNR-aware branch) would be fused with the compressed features to generate enhanced features. Finally, the enhanced image can
be obtained by decoding the enhanced features directly.
The experiments show that Our proposed joint solution surpasses sequential solutions significantly in terms of PSNR and MS-SSIM, resulting in superior reconstructed image quality for subsequent visual perception. Additionally, it offers lower computational costs and a reduced number of model parameters. 

\section{Acknowledgments}
This work was supported in part by the National Natural Science Foundation of China under Grant 62301228, 62176100, 62376011 and in part by the Special Project of Science and Technology Development of Central Guiding Local of Hubei Province under Grant 2021BEE056.
The computation is completed in the HPC Platform of Huazhong University of Science and Technology.
\bigskip
\bibliography{aaai24}
\clearpage
\newpage
\appendix
\section*{Summary}
This supplementary material is organized as follows.
\begin{itemize}
\item Section~\ref{section_1} introduces the architectures of the ``Joint Guidance without SNR-aware" and ``Joint Guidance with SNR-aware", and their training details.
\item Section~\ref{section_2} provides more experimental results about the impact of different branches on RD performance.
\item Section~\ref{section_3} provides more visualization results.
\end{itemize}

\section{Network Architecture and Training Details}
\label{section_1}
\subsection{Joint Guidance without SNR-aware.}
\begin{figure*}[!]
    \centering
	\includegraphics[width=\linewidth]{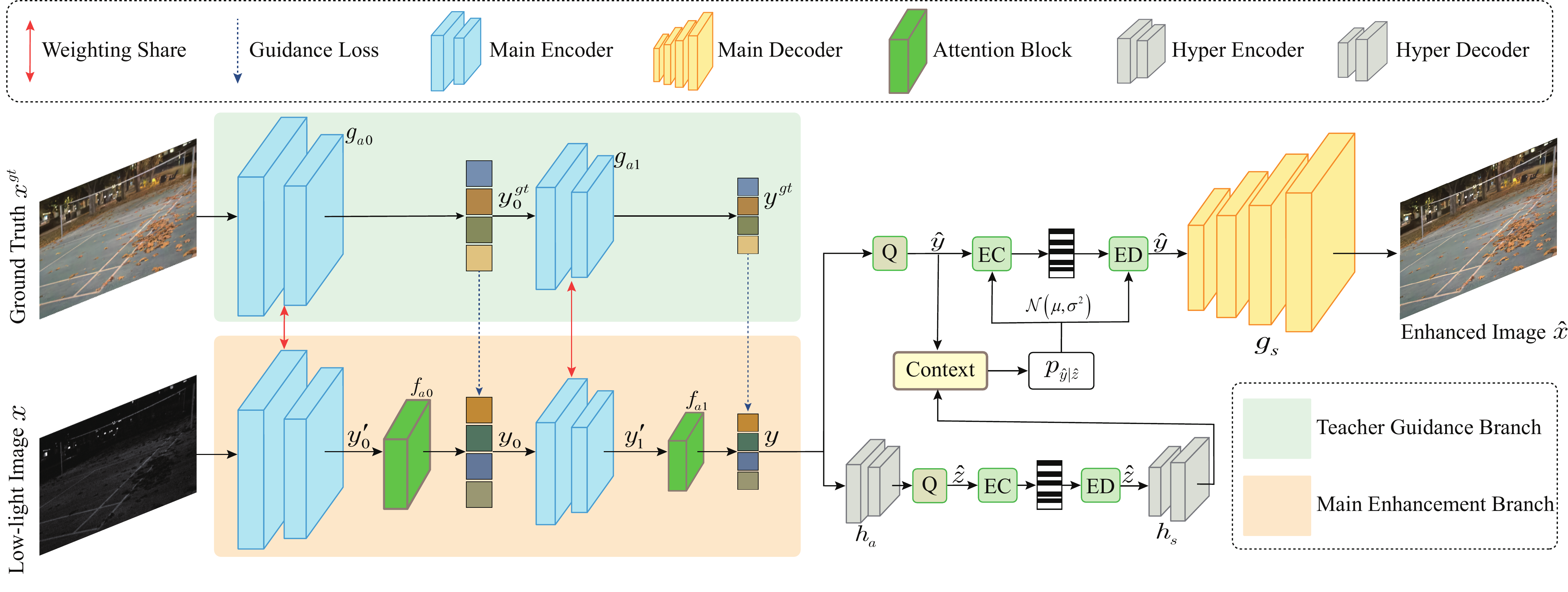}
	\caption{The Network architecture of the ``Joint Guidance without SNR-aware" is similar to DC~\cite{cheng2022optimizing}. The left half of the figure contains two branches, ``Teacher Guidance Branch" and ``Main Enhancement Branch". The right half of the figure contains the main decoder, entropy models, context model, and hyper encoder/decoder commonly used in learning-based compression methods~\cite{cheng2020learned, minnen2018joint}. Note that the ``Teacher Guidance Branch" is for training only and that the ``Main Enhancement Branch" and ``Attention Block" modules are activated during training of the entire network and used for inference.}
	\label{Joint_without_SNR_aware_architecture}
\end{figure*}

\paragraph{Network Architecture.}
The network architecture of ``Joint Guidance without SNR-aware" is shown in Figure~\ref{Joint_without_SNR_aware_architecture}. 
During the training procedure, the ground truth image $x^{gt}$ goes through the ``Teacher Guidance Branch" for the two-level guiding features~($y_{0}^{gt}$/$y^{gt}$).
The ``Teacher Guidance Branch" consists of main encoders~($g_{a0}$/$g_{a1}$). It provides guidance latent representations~($y_{0}^{gt}$/$y^{gt}$) which effectively supervise learning enhanced features.
The low-light image is fed into the ``Main Enhancement Branch" to obtain the two-level enhanced features~($y_{0}$/$y$).
The low-light features~($y_{0}^{\prime}$/$y_{1}^{\prime}$) are enhanced by ``Attention Block" modules~($f_{a0}$/$f_{a1}$). Finally, the enhanced image $\hat{x}$ is reconstructed by the main decoder $g_{s}$ directly.

\begin{figure*}[!]
    \centering
	\includegraphics[width=\linewidth]{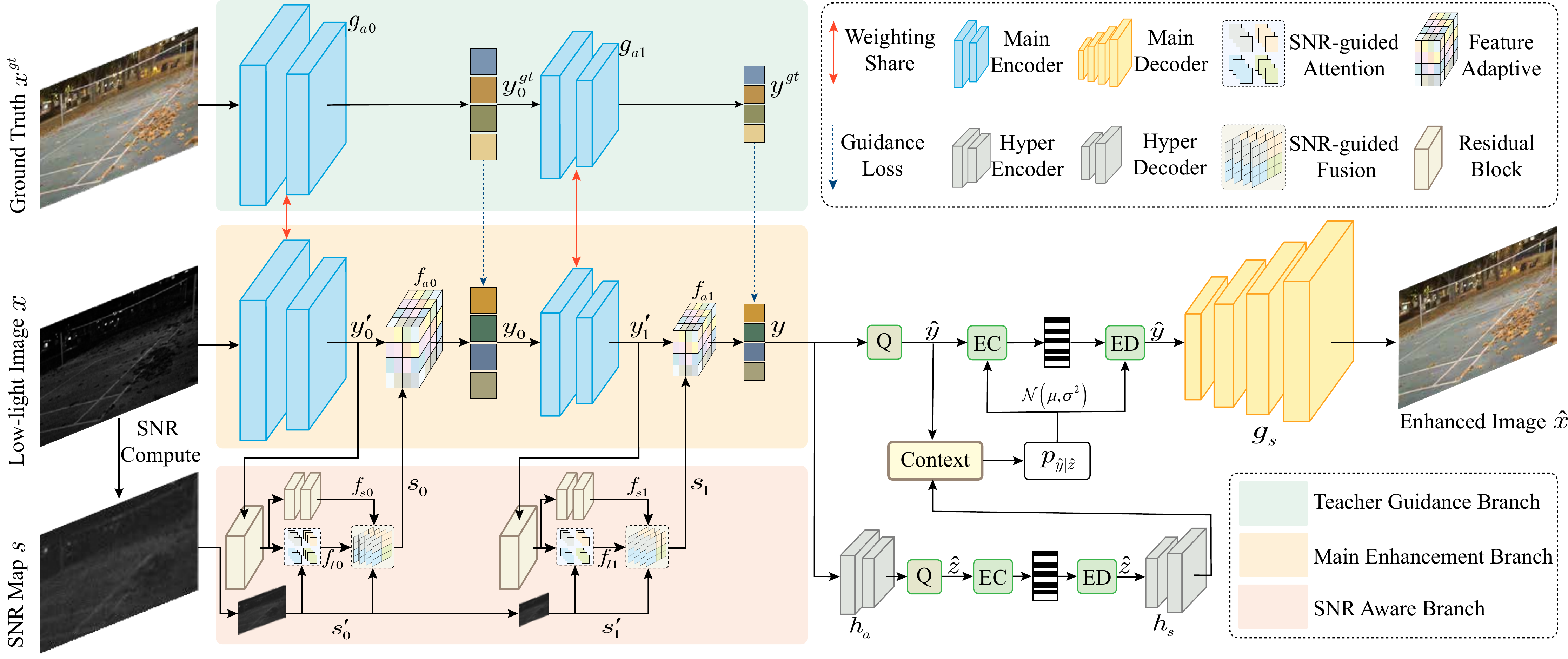}
	\caption{The Network architecture of the ``Joint Guidance with SNR-aware". The architecture contains three branches, ``Teacher Guidance Branch", ``Main Enhancement Branch" and ``SNR-Aware Branch", in the left half of the figure. The right half of the figure contains the main decoder, entropy models, context model, and hyper encoder/decoder commonly used in recent learning-based compression methods~\cite{cheng2020learned, minnen2018joint}. Note that the ``Teacher Guidance Branch" is for training only and that the ``Main Enhancement Branch" and ``SNR Aware Branch" are activated during training of the entire network and used for inference.}
\label{Joint_share_SNR_aware_architecture} 
\end{figure*}

\paragraph{Training Details.}
\label{train_details}
In our experiments, we observe that training both image compression and low-light enhancement tasks jointly at the beginning results in convergence problems. Thus, we adopt the two-stage training.
\par
We pre-train the framework without joining the ``Teacher Guidance Branch". In this case, the network architecture~(except for ``Attention Block" modules) is similar to the Cheng2020-anchor model of the CompressAI library~\cite{begaint2020compressai} implementation. The optimization loss can be temporarily overwritten as:
\begin{small}
\begin{equation}
\begin{aligned}
	 \mathcal{L}=&\lambda_{d} \cdot \mathcal{D}(x,\hat{x}) + \mathcal{R}(\hat{y}) + \mathcal{R}(\hat{z}) \\
	 =&\lambda_{d} \cdot \mathbb{E}_{x\sim p_{x}}\bigl[\left\|x - \hat{x}\right\|_{\mathfrak{p}}^{\mathfrak{p}}\bigr]\\ 
	 & - \mathbb{E}_{\hat{y} \sim q_{\hat{y}}}\bigl[\log p_{\hat{y}|\hat{z}}(\hat{y}|\hat{z})\bigr] - \mathbb{E}_{\hat{z} \sim q_{\hat{z}}}\bigl[\log p_{\hat{z}}(\hat{z})\bigr].
\label{Joint_loss_pre}
\end{aligned}
\end{equation}
\end{small}
Where $x$ and $\hat{x}$ denote the original image and decoded image respectively. We set the $\lambda_{d}=0.0016$. It is worth noting that the parameter $\mathfrak{p}$ of  the first term $\mathbb{E}_{x\sim p_{x}}\bigl [\left\|x - \hat{x}\right\|_{\mathfrak{p}}^{\mathfrak{p}}\bigr ]$ is equal to 2. That means, the distortion loss $\mathcal{D}(x,\hat{x})$ is the MSE loss instead of $L_{1}$ loss.
\par
We train the entire network by loading the pre-trained parameters. The joint optimization loss is Equation~\ref{Joint_loss}. The $\lambda_{d}$ and $\lambda_{g}$ in the Equation~\ref{Joint_loss} are tuned with fixed ratio~($\frac{\lambda_{d}}{\lambda_{g}}=const$) to get various compression rates. The parameter $\mathfrak{p}$ of  the first term $\mathbb{E}_{x\sim p_{x}}\bigl [\left\|x^{gt} - \hat{x}\right\|_{\mathfrak{p}}^{\mathfrak{p}}\bigr ]$ is equal to 1.
\begin{small}
\begin{equation}
\begin{aligned}
	 \mathcal{L}&=\lambda_{d} \cdot \mathcal{D}(x^{gt},\hat{x}) + \lambda_{g} \cdot \mathcal{S}(y_{0}^{gt},y_{0},y^{gt},y) + \mathcal{R}(\hat{y}) + \mathcal{R}(\hat{z}) \\
	                     &=\lambda_{d} \cdot \mathbb{E}_{x\sim p_{x}}\bigl [\left\|x^{gt} - \hat{x}\right\|_{\mathfrak{p}}^{\mathfrak{p}}\bigr ] + \lambda_{g} \cdot \mathbb{E}_{y_{0}, y \sim q}\bigl [\left\| y^{gt}_{0}- y_{0} \right\|_{1} + \\
	                      & \left\| y^{gt}- y \right\|_{1}\bigr ]  - \mathbb{E}_{\hat{y} \sim q_{\hat{y}}}\bigl [\log p_{\hat{y}|\hat{z}}(\hat{y}|\hat{z})\bigr ] - \mathbb{E}_{\hat{z} \sim q_{\hat{z}}}\bigl [\log p_{\hat{z}}(\hat{z})\bigr ].
\label{Joint_loss}
\end{aligned}
\end{equation}
\end{small}
The first term $\mathcal{D}(x^{gt},\hat{x})$ measures distortion between ground truth image $x^{gt}$ and reconstructed image $\hat{x}$. In our experiment, using $L_{1}$ distortion loss is more beneficial for the stability of training. The second term $ \mathcal{S}(y_{0}^{gt},y_{0},y^{gt},y)$ measures the sum of two-level errors between ground truth latent representations $(y^{gt}_{0}/y^{gt})$ and corresponding enhanced latent representations $(y_{0}/y)$ by using $L_{1}$ distortion loss. The third term $\mathcal{R}(\hat{y})$ and forth term $ \mathcal{R}(\hat{z})$ denote compression levels. $\lambda_{d}$ and $\lambda_{g}$ denote the weighting coefficients, which are the trade-off between compression levels and distortion.

\subsection{Joint Guidance with SNR-aware.}
\paragraph{Network Architecture.}
We additionally use a three-branch network architecture named ``Joint Guidance and SNR-aware''. The architecture is shown in Figure~\ref{Joint_share_SNR_aware_architecture}.
During the training stage, the ground truth image $x^{gt}$ goes through the ``Teacher Guidance Branch" for the two-level guiding features~($y_{0}^{gt}$/$y^{gt}$).
The SNR map $s$ is achieved by employing a no-learning-based denoising operation which is simple yet effective. Local and non-local information on the low-light image is obtained through the ``SNR Aware Branch". The low-light features~($y_{0}^{\prime}$ / $y_{1}^{\prime}$) combine with the local and non-local information~($s_{0}$/$s_{1}$) generated by the ``SNR Aware Branch" to obtain the enhanced latent representations~($y_{0}$/$y$). Finally, the enhanced features $\hat{y}$ are fed into the main decoder $g_{s}$ to obtain the enhanced image $\hat{x}$.
\paragraph{Training Details.}
The training details are similar to the ``Joint Guidance without SNR-aware" method, please refer to ``Training Details.'' in Section~\ref{train_details}. It is worth noting that using a three-branch architecture for training is costly.

\begin{figure*}[h]
\centering
\subcaptionbox{PSNR on SDSD-indoor} {\includegraphics[width = .245\linewidth]{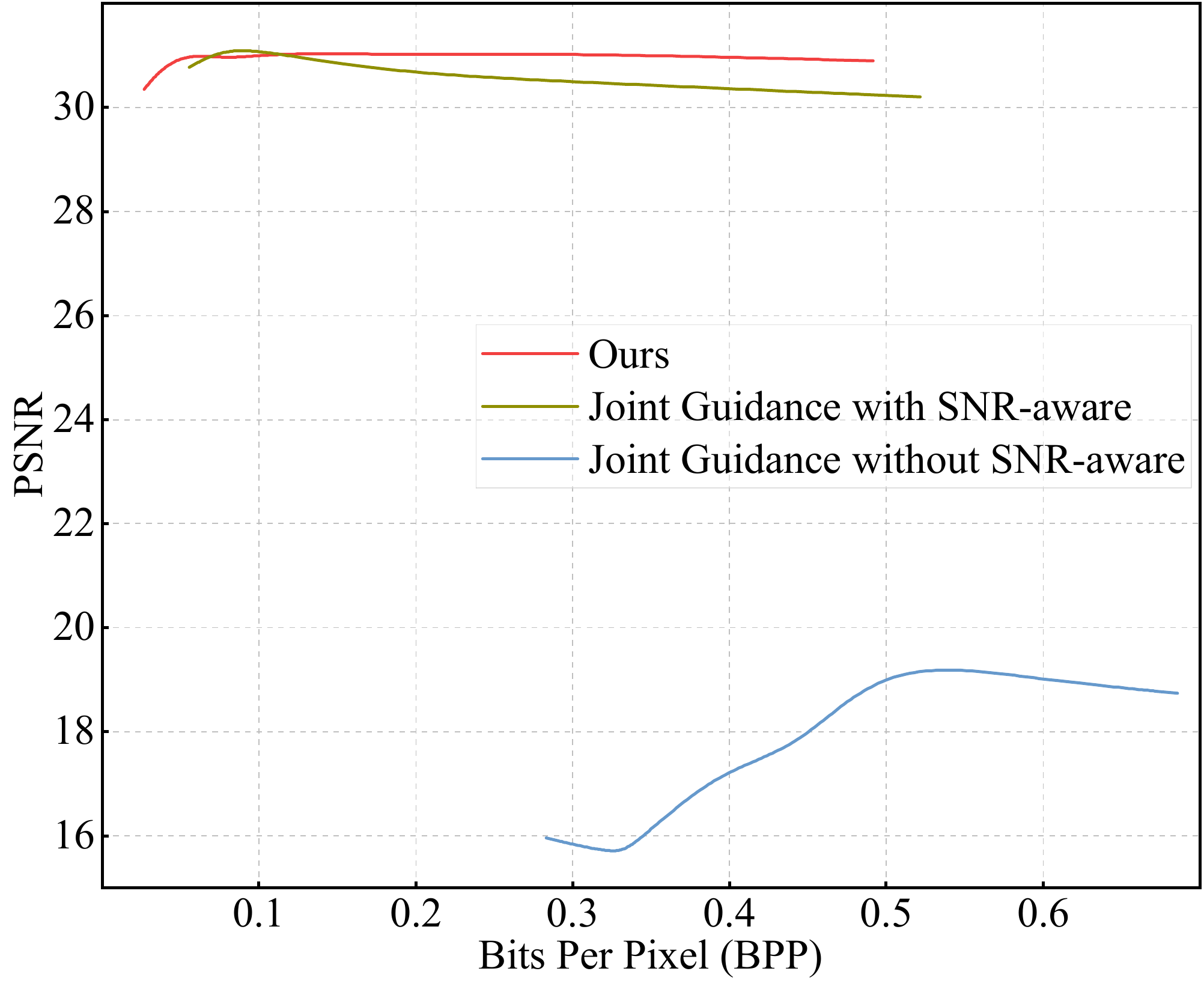}}\hfill
\subcaptionbox{MS-SSIM on SDSD-indoor}{\includegraphics[width = .245\linewidth]{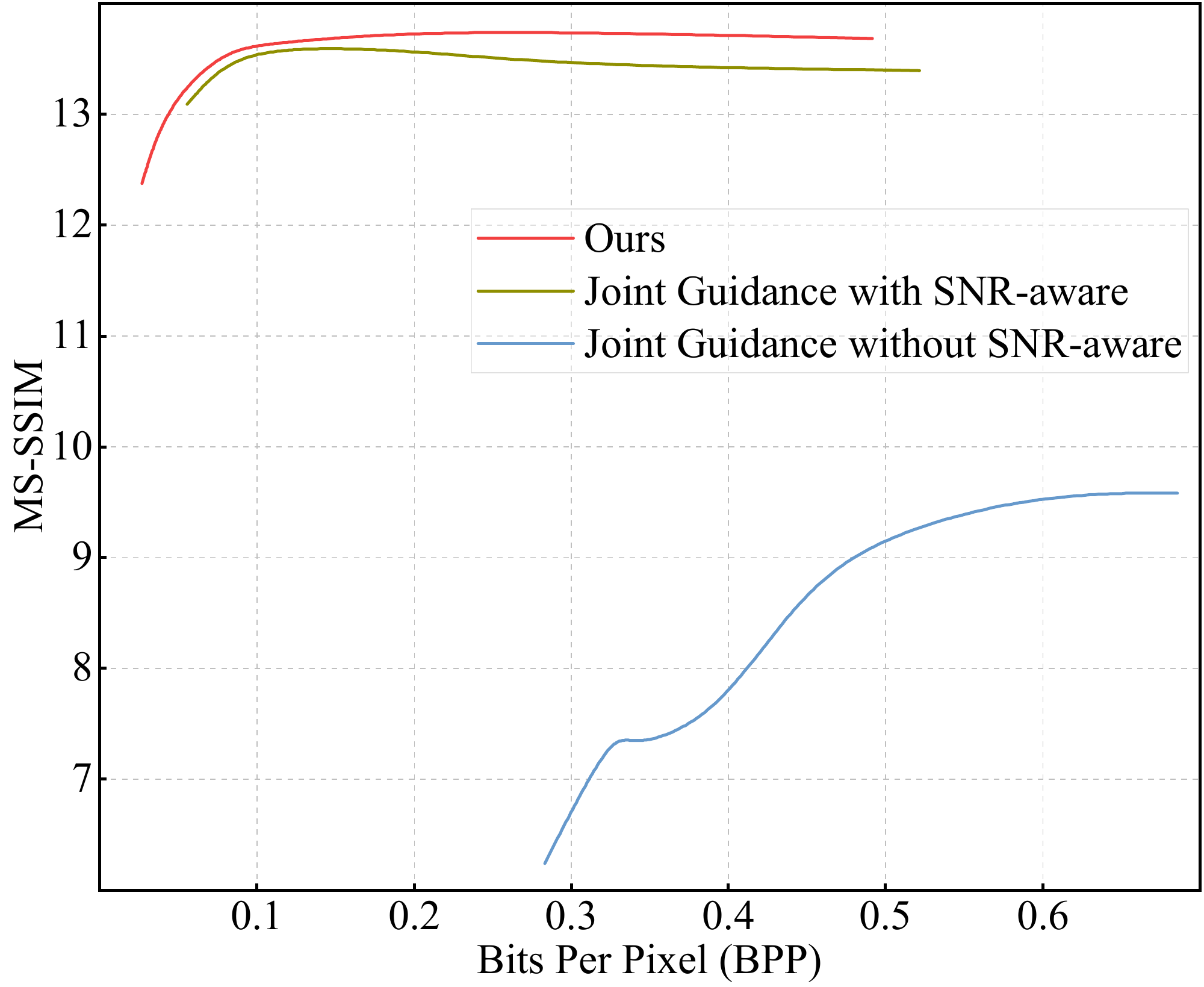}}\hfill
\subcaptionbox{PSNR on SDSD-outdoor}{\includegraphics[width = .245\linewidth]{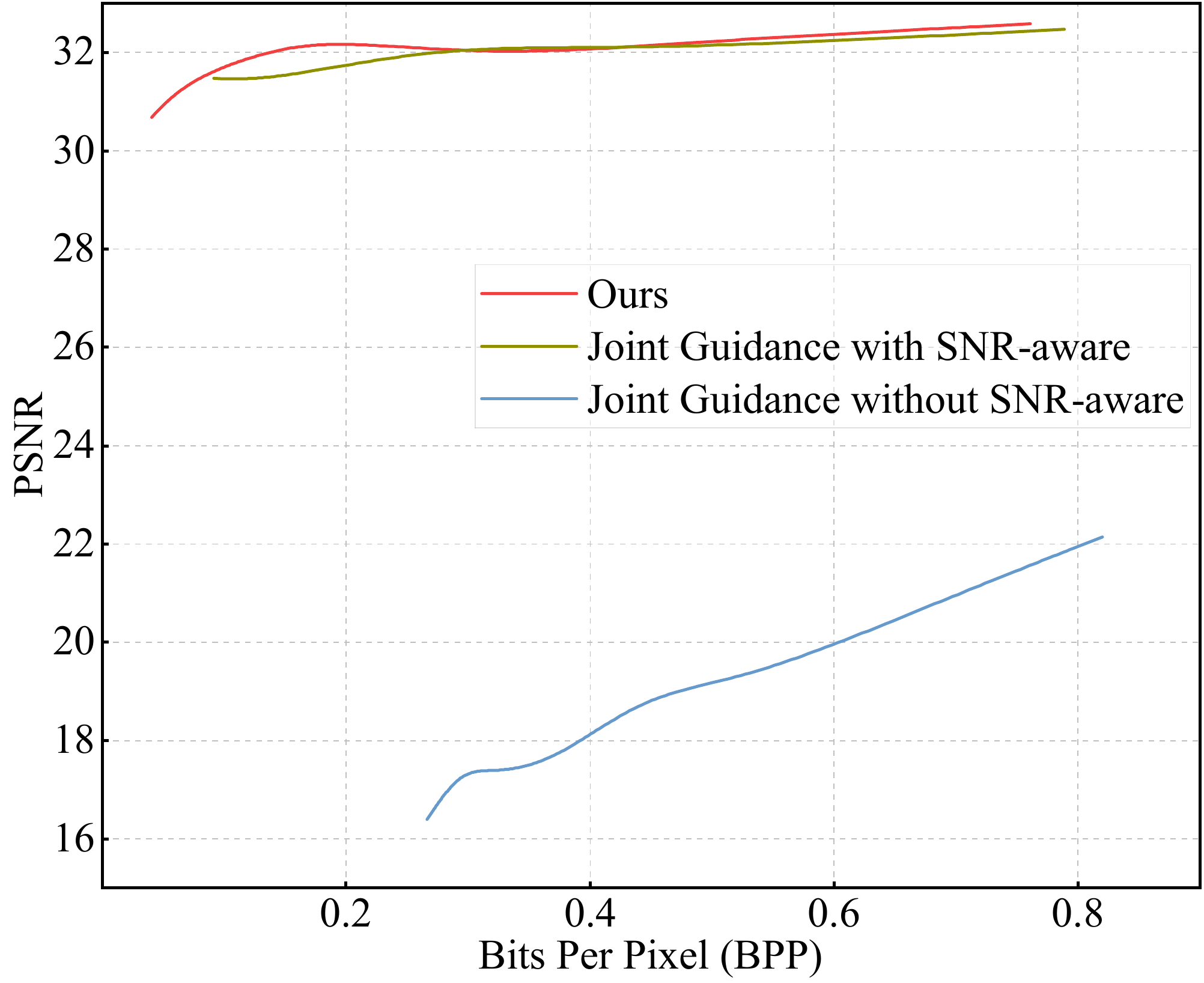}}\hfill
\subcaptionbox{MS-SSIM on SDSD-outdoor}{\includegraphics[width = .245\linewidth]{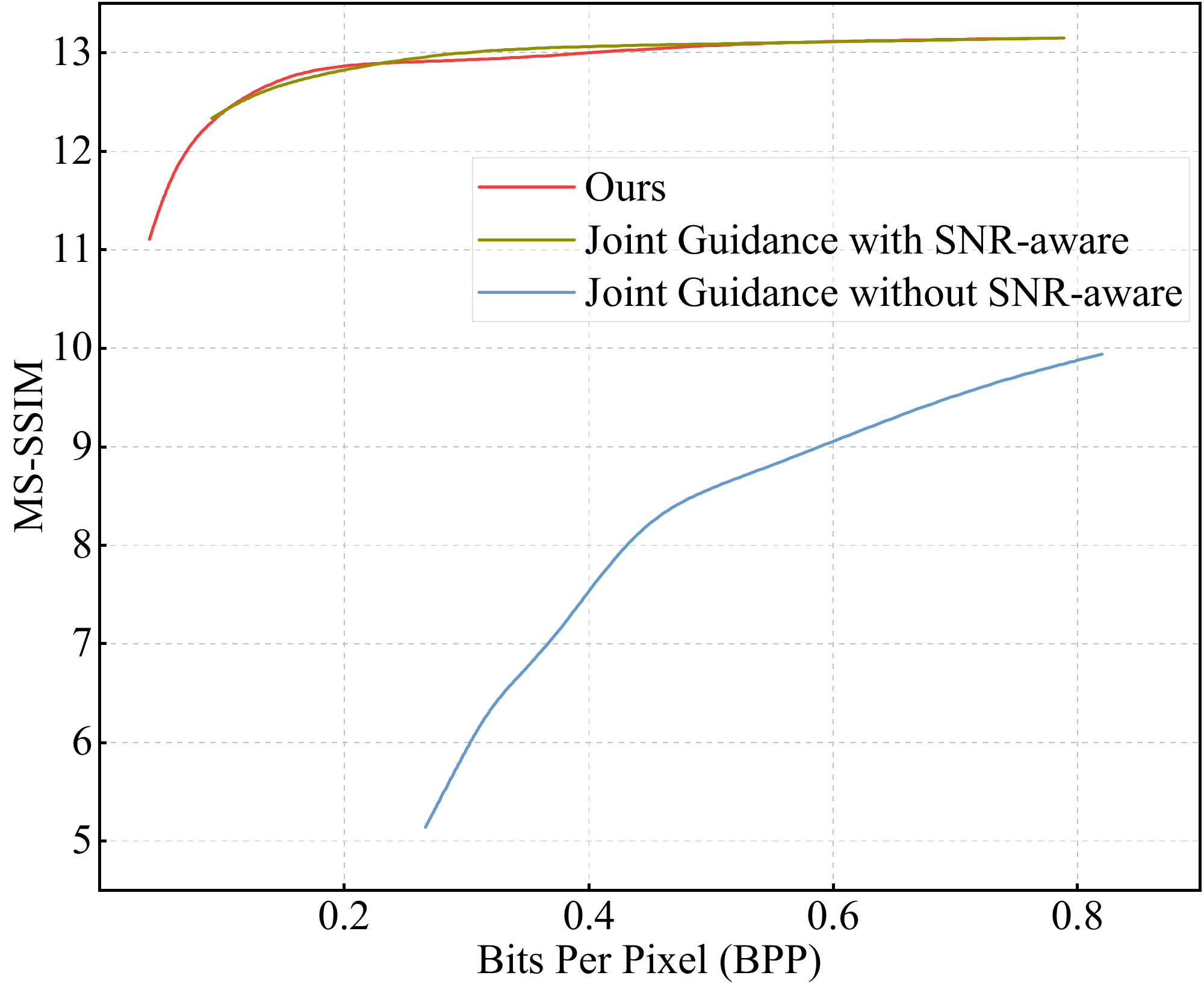}}
\caption{RD performance curves aggregated over two datasets. (a)/(c) and (b)/(d) are results on SDSD-indoor and SDSD-outdoor datasets about PSNR and MS-SSIM, respectively.}
\label{RD_GW_curve_additional}
\end{figure*}

\begin{figure*}[h]
	\centering
	\includegraphics[width=\linewidth, height=8.3cm]{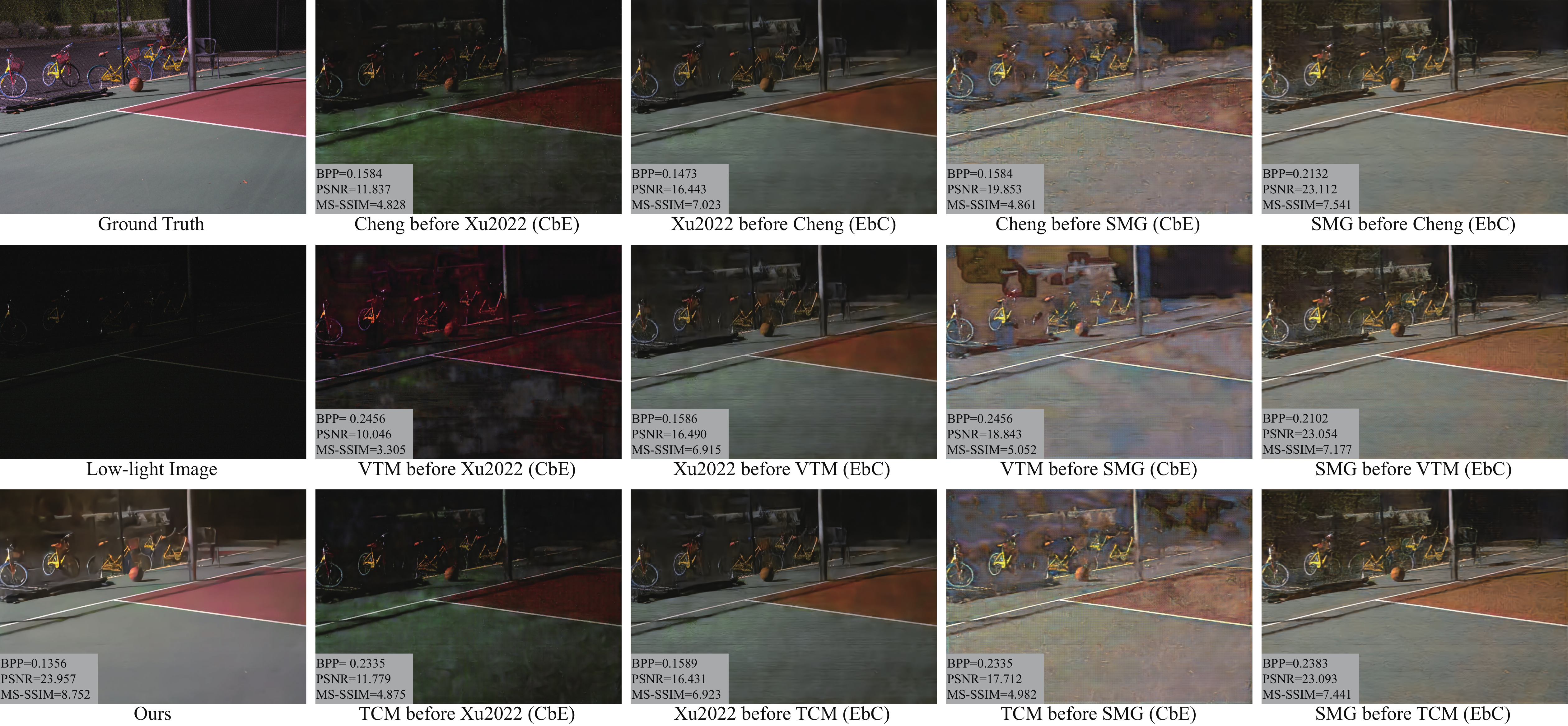}
	\label{visual_results_figure1}
\end{figure*}

\section{More Analyze Experiments}
\label{section_2}
The experimental results of the ``Joint Guidance without SNR-aware'' and the ``Joint Guidance with SNR-aware'' methods on SDSD-indoor and SDSD-outdoor datasets are shown in Figure~\ref{RD_GW_curve_additional}. 
Compared with the ``Joint Guidance without SNR-aware'', our proposed joint solution is more effective than ``Joint Guidance without SNR-aware".
This method of simply using corresponding network modules in the main enhancement branch is ineffective for joint image compression and low-light enhancement tasks.
The results show that our proposed solution outperforms the ``Joint Guidance without SNR-aware" by a large margin, indicating the significance and importance of the SNR-aware branch~(red curve vs blue curve in Figure~\ref{RD_GW_curve_additional}).
In addition, the performance of using such a ``Teacher Guidance Branch" is slightly worse than our joint solution~(red curve vs yellow curve in Figure~\ref{RD_GW_curve_additional}), while additionally increasing the computational cost during the training procedure. Our usage of SNR-aware information is more effective and efficient.

\section{More Visualization Results}
\label{section_3}
The proposed joint solution compares with the twelve sequential solutions as follows:
(1) ``Cheng before Xu2022 (CbE)'';
(2) ``VTM before Xu2022 (CbE)'';
(3) ``TCM before Xu2022 (CbE)'';
(4) ``Xu2022 before Cheng (EbC)'';
(5) ``Xu2022 before VTM (EbC)'';
(6) ``Xu2022 before TCM (EbC)'';
(7) ``Cheng before SMG (CbE)'';
(8) ``VTM before SMG (CbE)'';
(9) ``TCM before SMG (CbE)'';
(10) ``SMG before Cheng (EbC)'';
(11) ``SMG before VTM (EbC)'';
(12) ``SMG before TCM (EbC)''.
For brief representation, ``Cheng" denotes the compression method~\cite{cheng2020learned}, ``VTM" denotes the classical codec method VVC~\cite{vvc}, ``TCM'' denotes the compression method~\cite{liu2023learned}, ``Xu2022'' denotes the low-light enhancement method~\cite{xu2022snr}, ``SMG'' denotes the low-light enhancement method~\cite{xu2023low}.
Those results further indicate that our proposed joint solution can indeed alleviate the problem of error accumulation and loss of information in the individual models that plague the sequential solution.


\begin{figure*}[!]
	\centering
	\includegraphics[width=\linewidth, height=10cm]{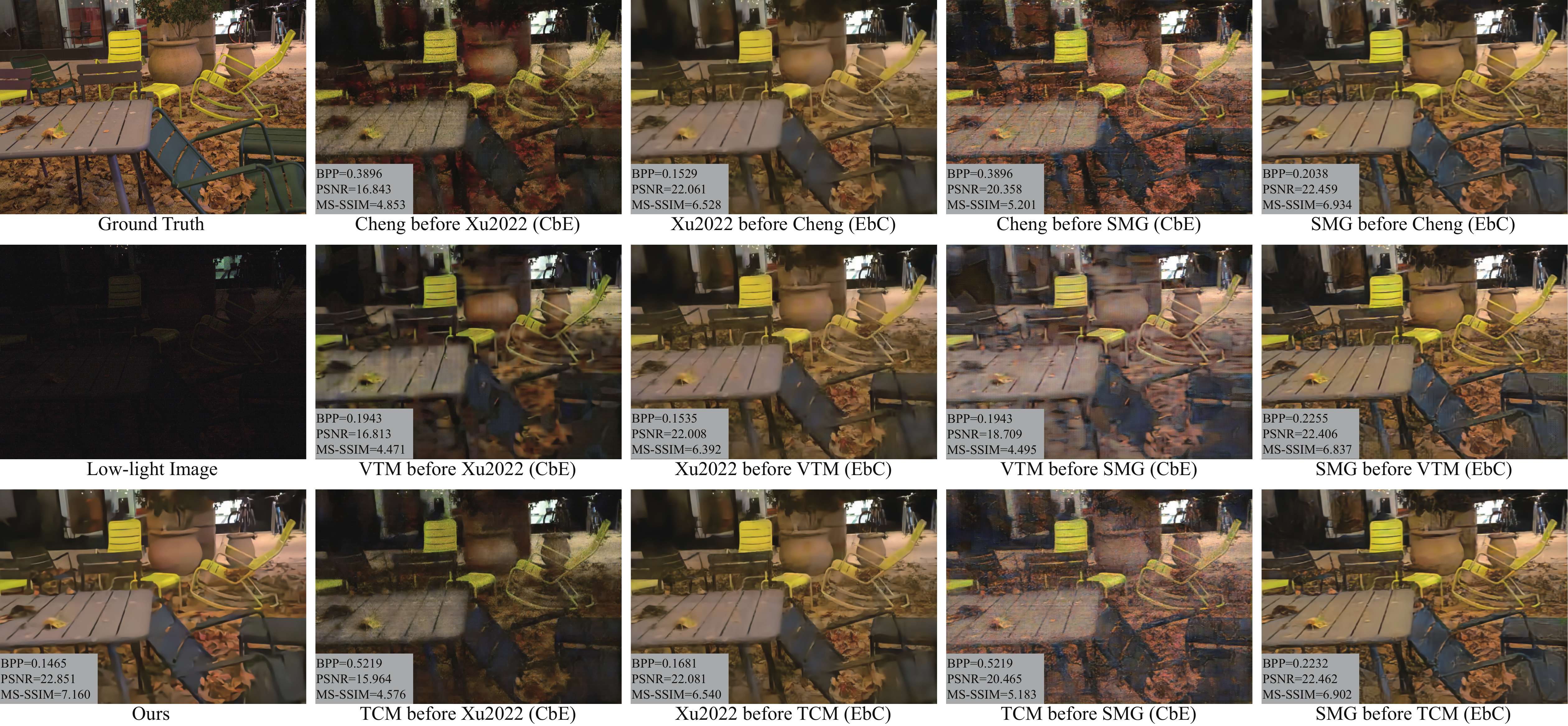}
	\label{visual_results_figure2}
\end{figure*}

\begin{figure*}[!]
	\centering
	\includegraphics[width=\linewidth, height=10cm]{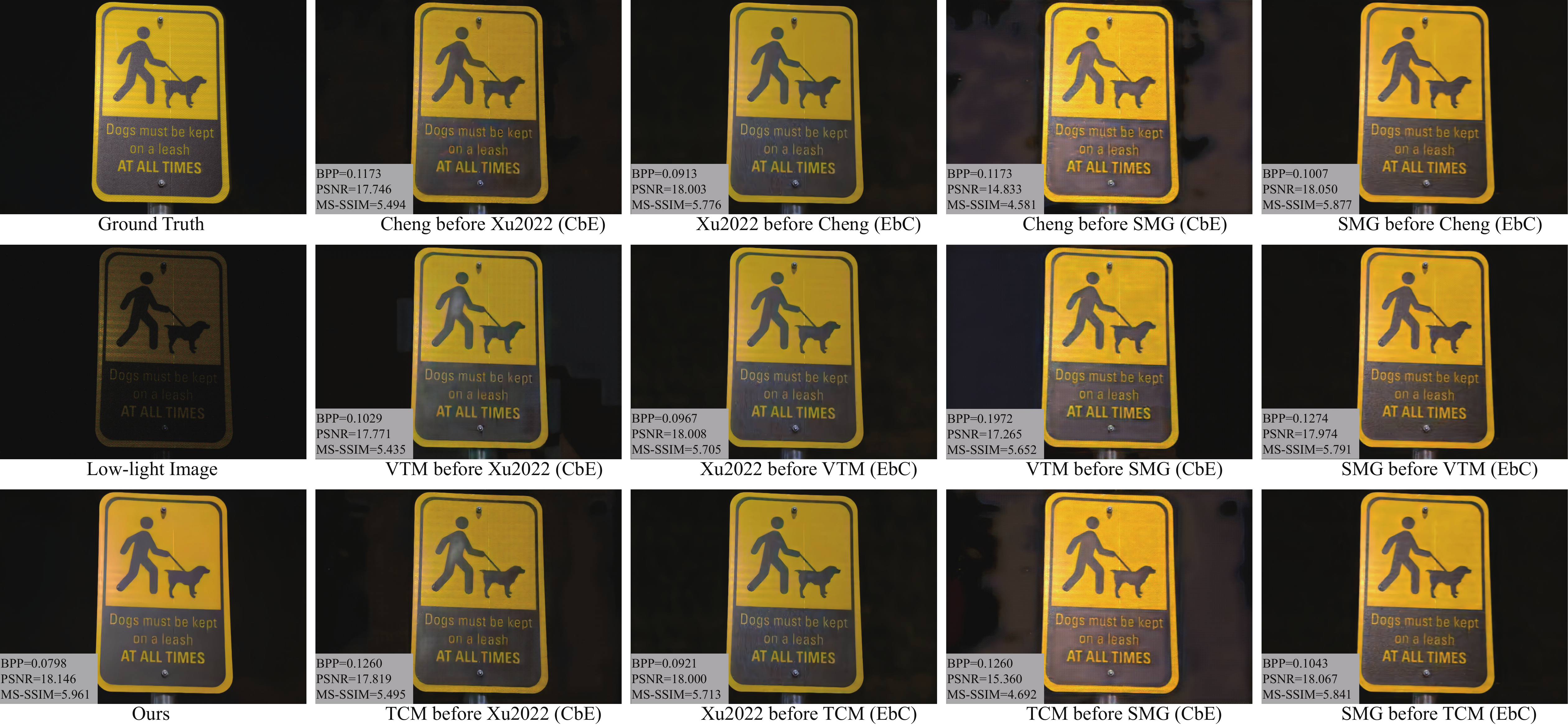}
	\label{visual_results_figure3}
\end{figure*}

\begin{figure*}[!]
	\centering
	\includegraphics[width=\linewidth, height=10cm]{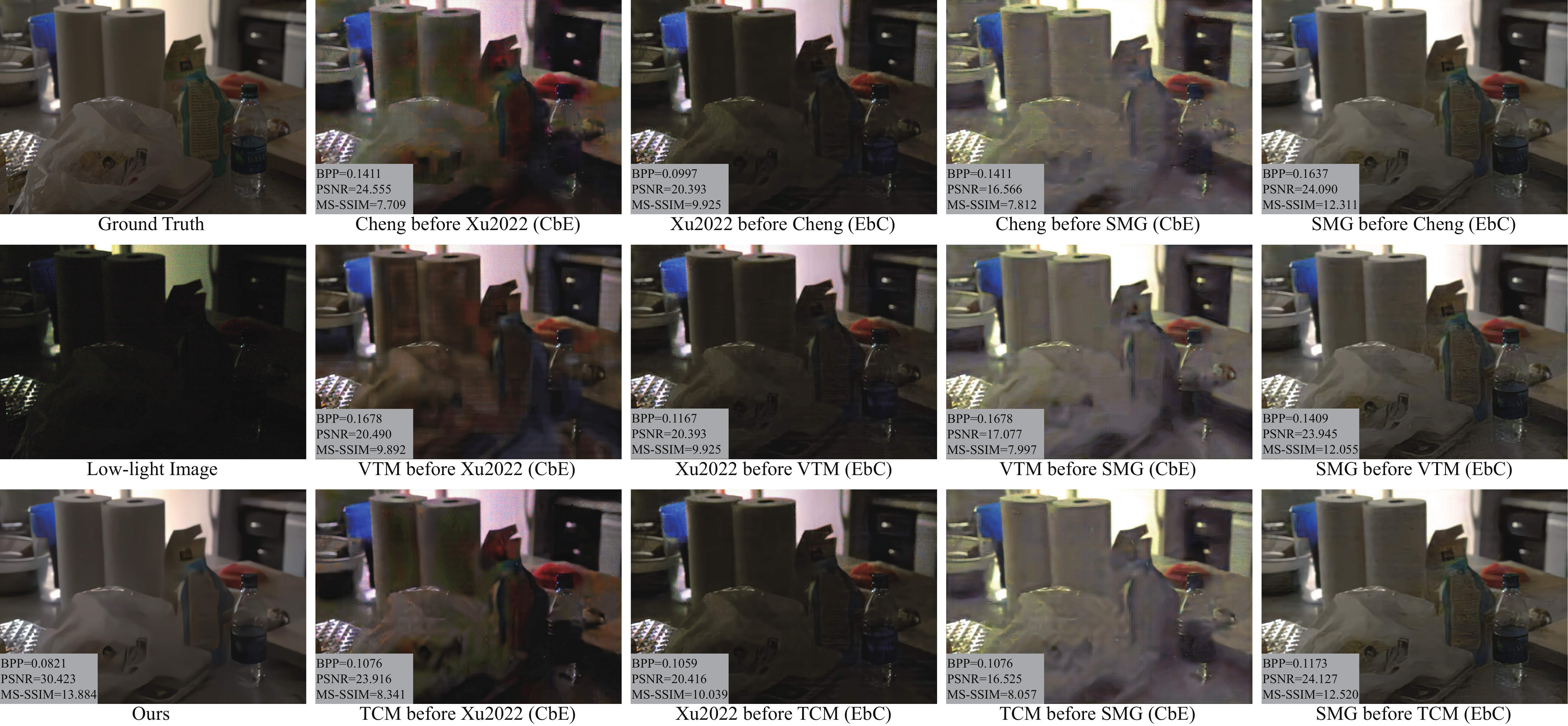}
	\label{visual_results_figure4}
\end{figure*}

\begin{figure*}[!]
	\centering
	\includegraphics[width=\linewidth, height=10cm]{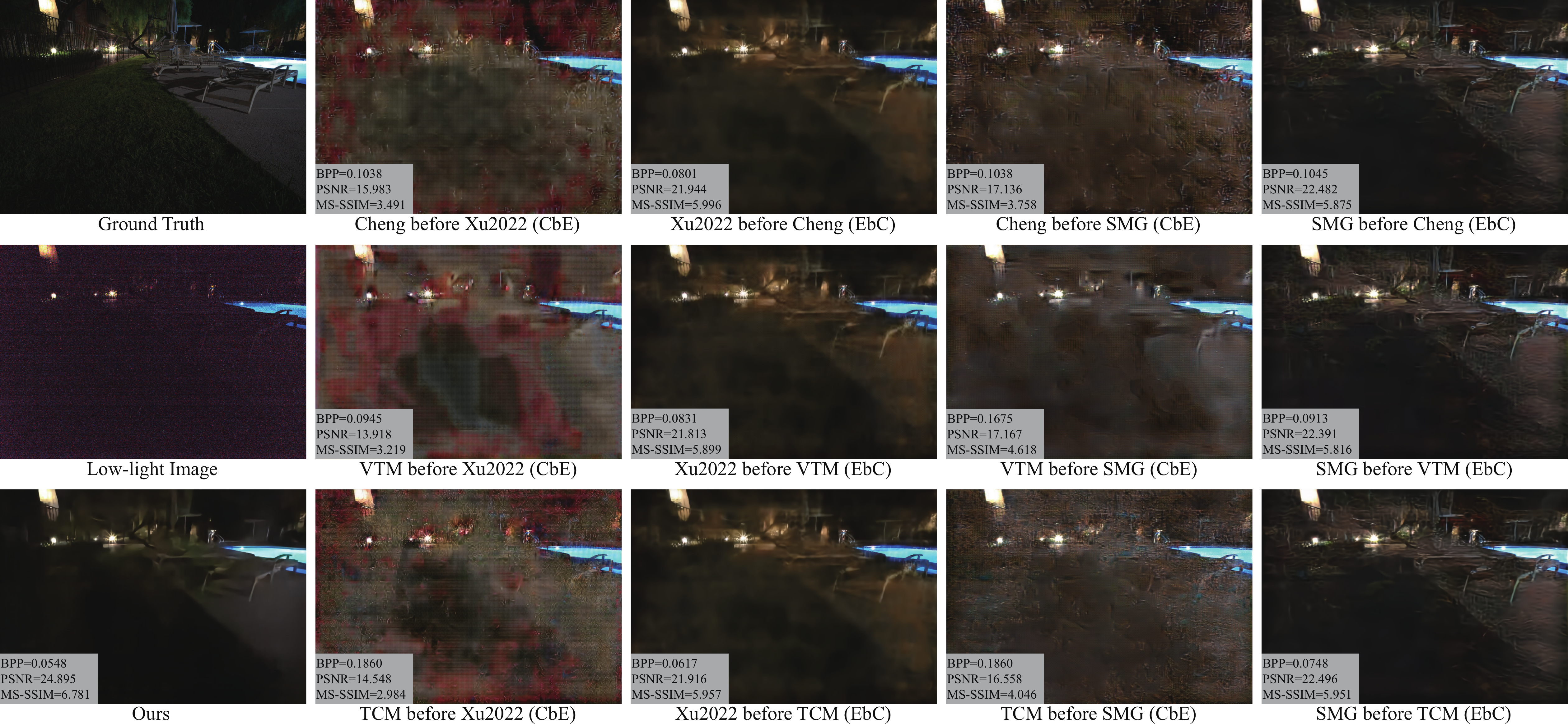}
	\label{visual_results_figure5}
\end{figure*}

\begin{figure*}[!]
	\centering
	\includegraphics[width=\linewidth, height=10cm]{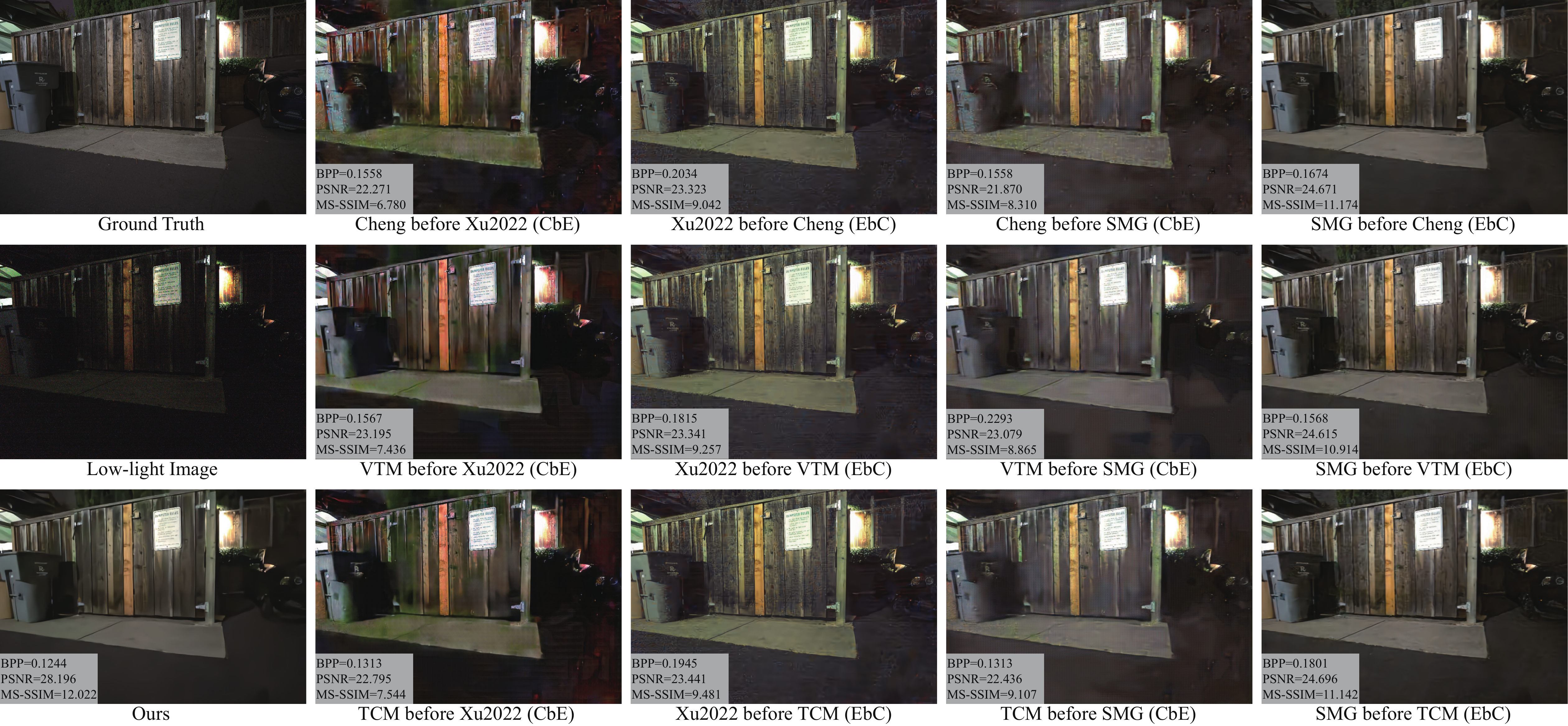}
	\label{visual_results_figure6}
\end{figure*}

\begin{figure*}[!]
	\centering
	\includegraphics[width=\linewidth, height=10cm]{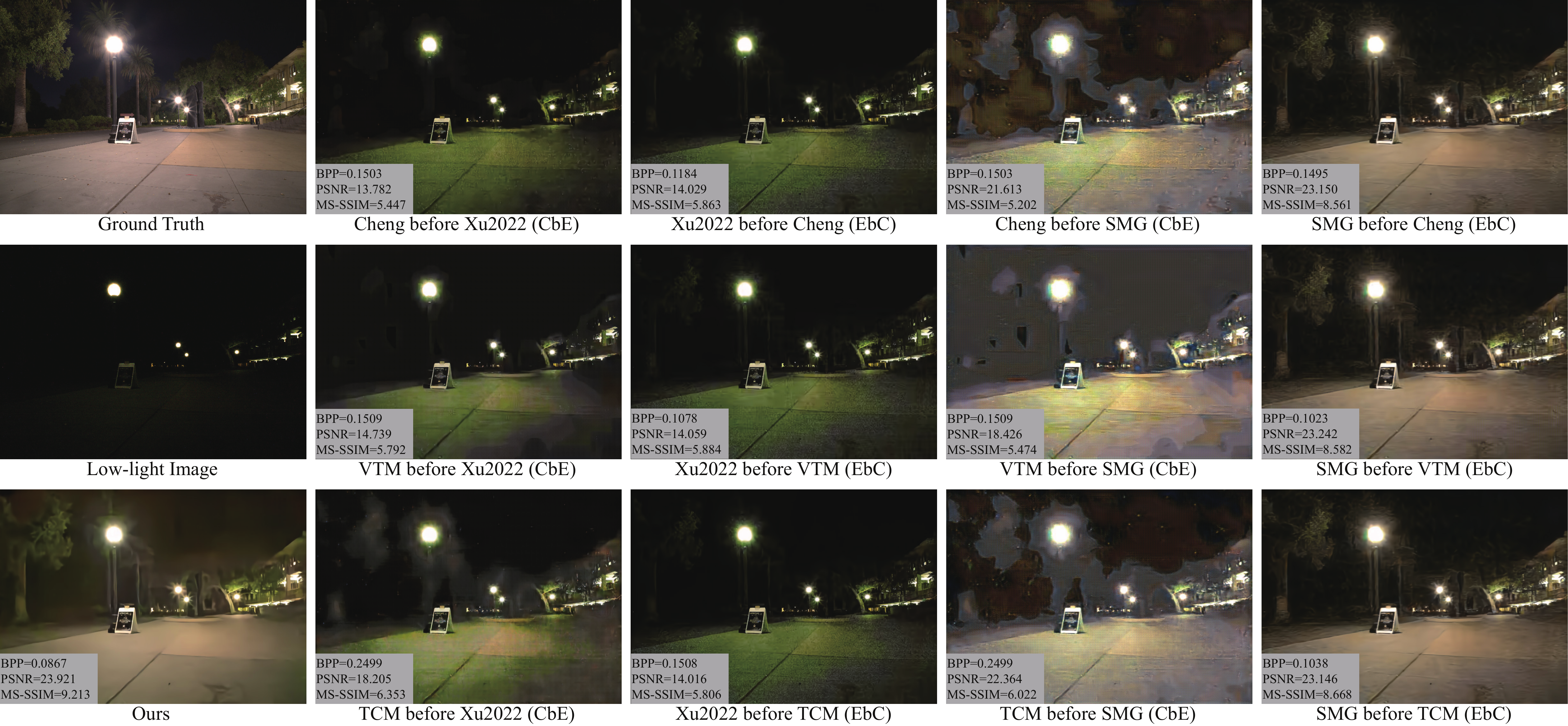}
	\label{visual_results_figure7}
\end{figure*}

\begin{figure*}[!]
	\centering
	\includegraphics[width=\linewidth, height=10cm]{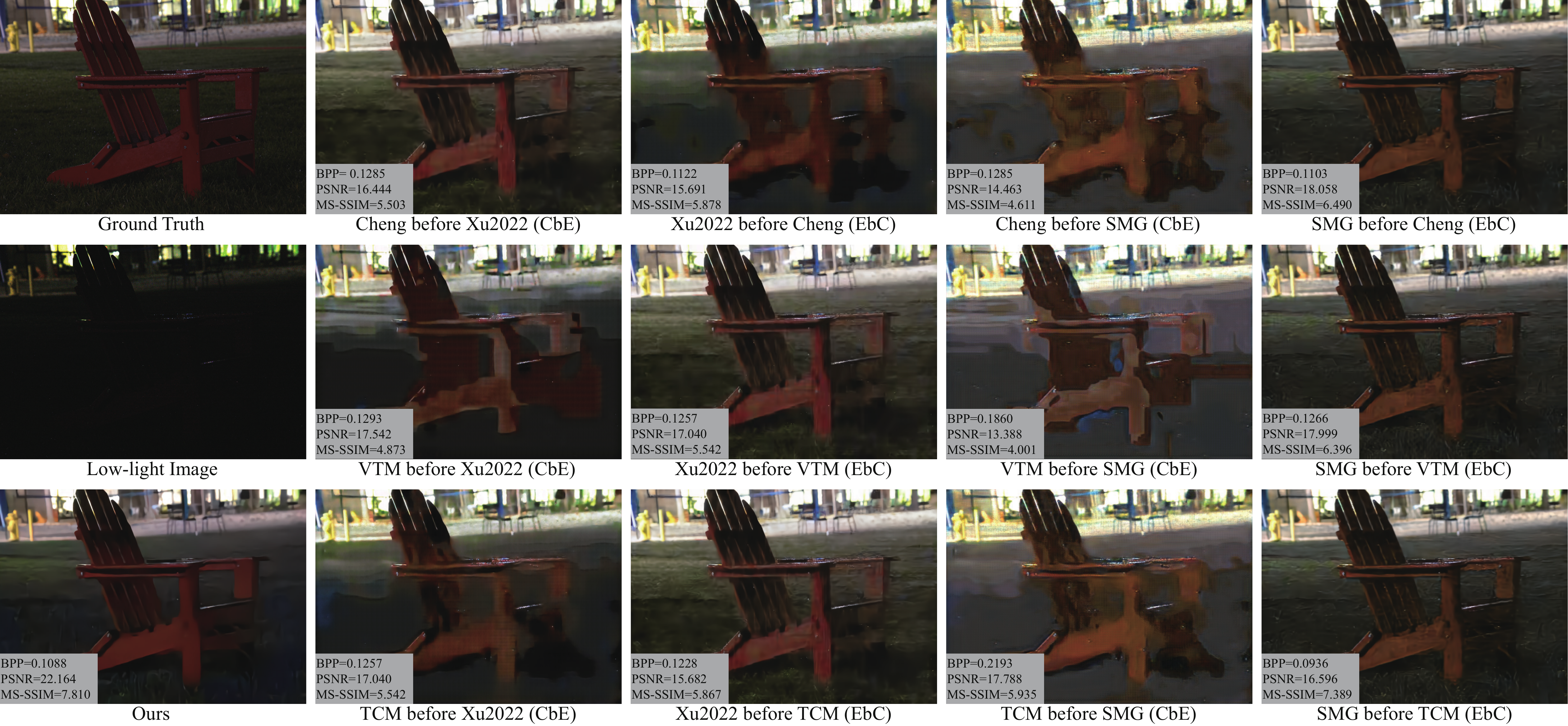}
	\label{visual_results_figure8}
\end{figure*}

\begin{figure*}[!]
	\centering
	\includegraphics[width=\linewidth, height=10cm]{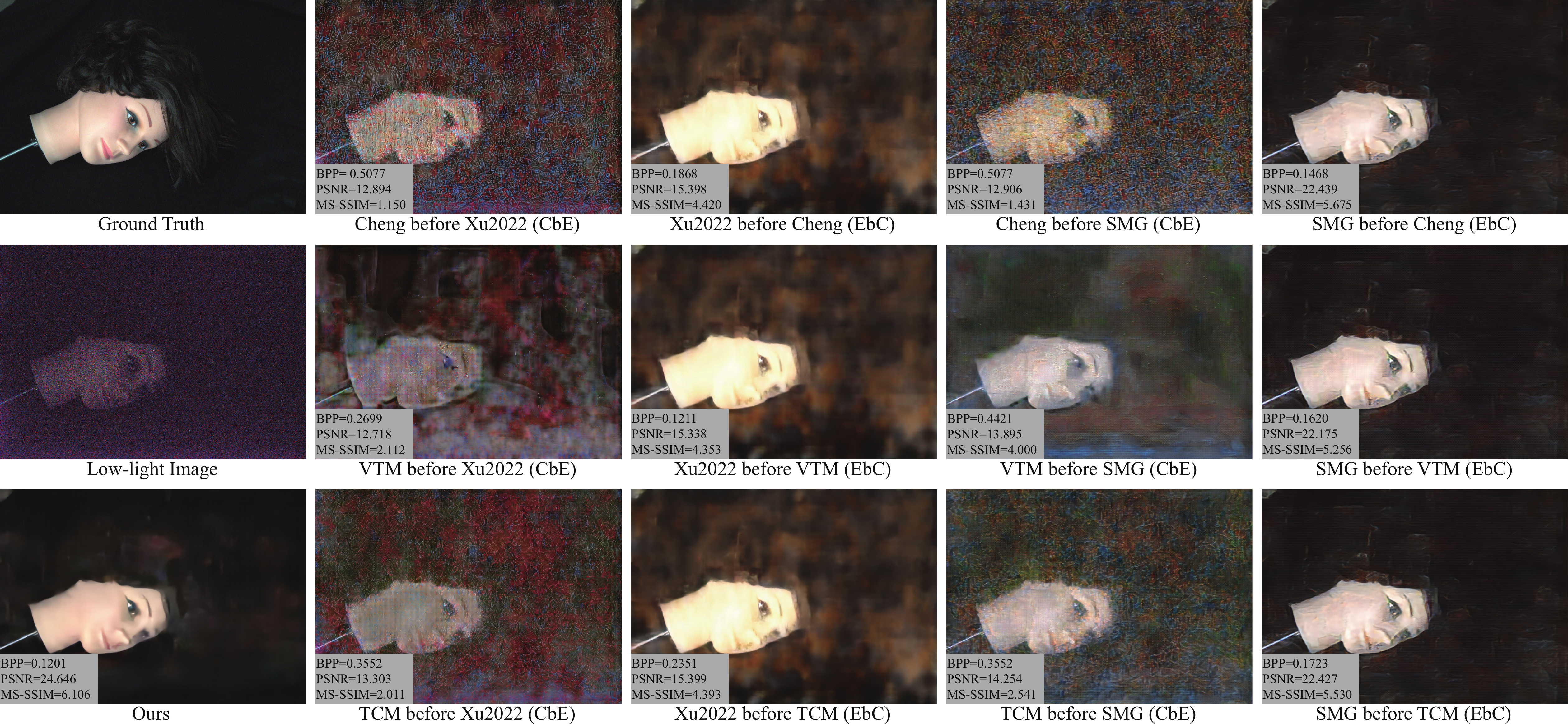}
	\label{visual_results_figure9}
\end{figure*}

\begin{figure*}[!]
	\centering
	\includegraphics[width=\linewidth, height=10cm]{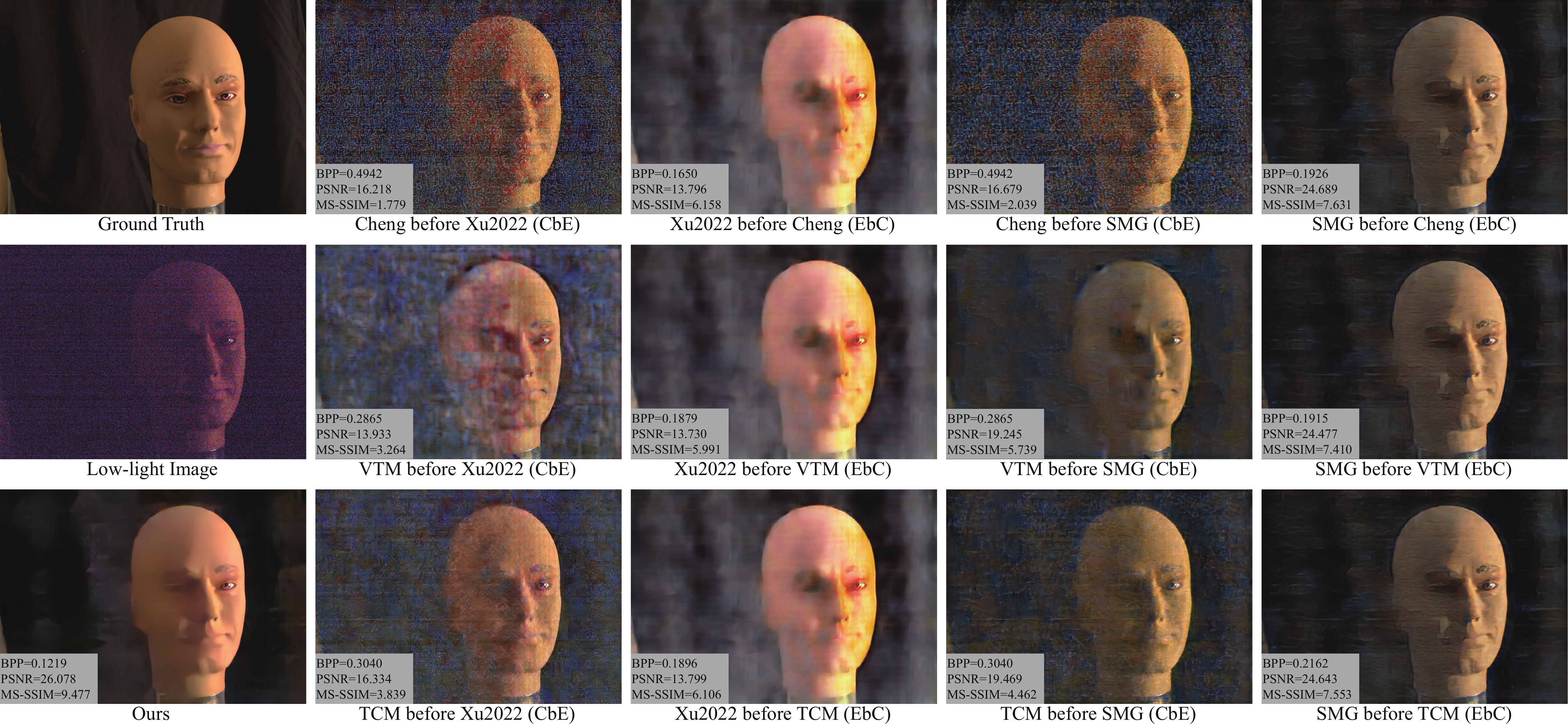}
	\label{visual_results_figure10}
\end{figure*}

\begin{figure*}[!]
	\centering
	\includegraphics[width=\linewidth, height=10cm]{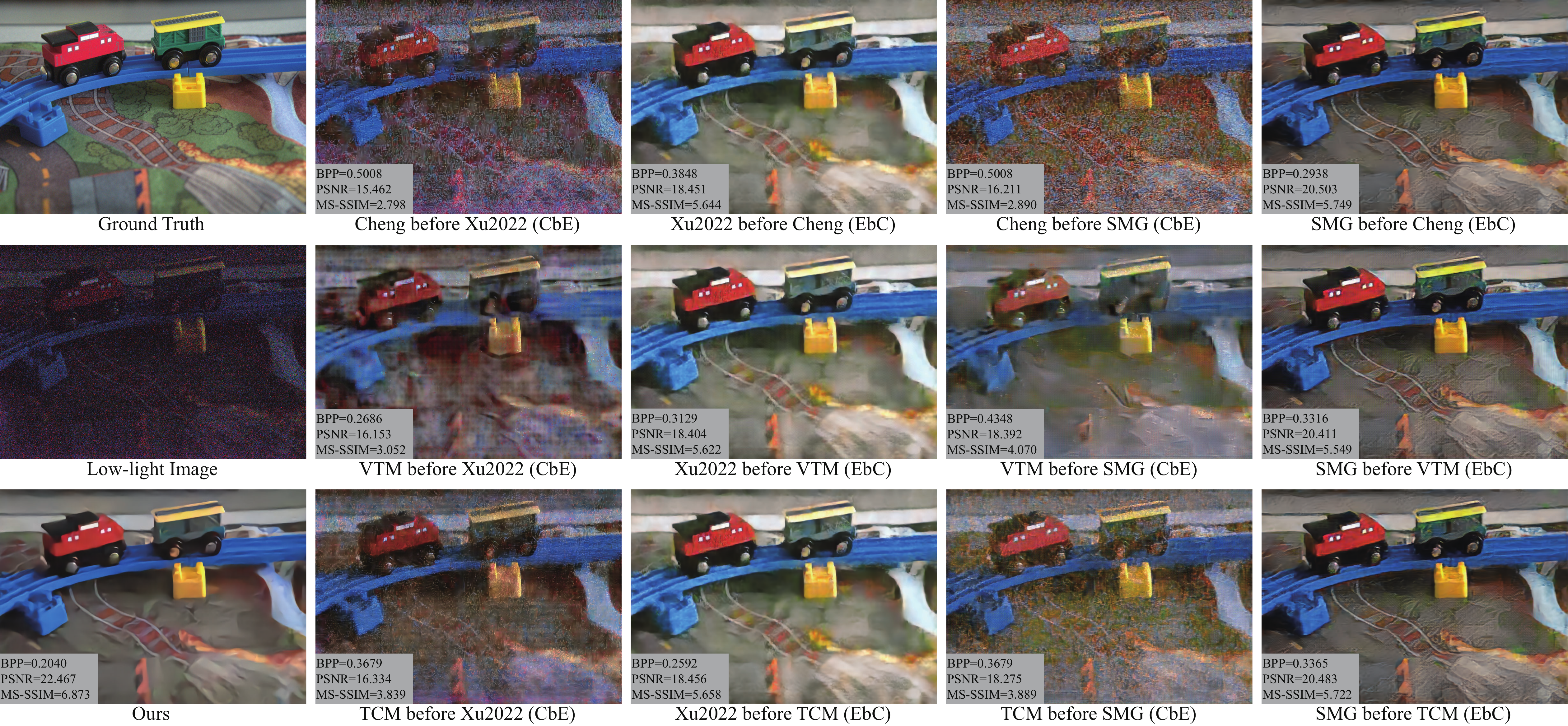}
	\label{visual_results_figure11}
\end{figure*}
\end{document}